\pdfoutput=1
\documentclass[11pt,a4paper]{article}

\usepackage[utf8]{inputenc}
\usepackage[T1]{fontenc}
\usepackage{graphicx}
\usepackage[margin=1in]{geometry}
\usepackage[authoryear]{natbib}
\usepackage{amssymb}
\usepackage{booktabs}
\usepackage{array}
\usepackage{threeparttable}
\usepackage{float}
\usepackage{tikz}
\usepackage{hyperref}
\usepackage[skins]{tcolorbox}
\usepackage{caption}
\hypersetup{
  colorlinks=true,
  allcolors=black
}

\title{\textbf{The role of spatial context and multitask learning in the detection of organic and conventional farming systems based on Sentinel-2 time series}}

\author{
Jan Hemmerling\textsuperscript{1,2,*},
Marcel Schwieder\textsuperscript{1,2},
Philippe Rufin\textsuperscript{3,2},\\
Leon-Friedrich Thomas\textsuperscript{4},
Mirela Tulbure\textsuperscript{5},
Patrick Hostert\textsuperscript{2},
Stefan Erasmi\textsuperscript{1}
}

\date{} 

\begin{document}
\maketitle

\begin{flushleft}
\textsuperscript{1} Thünen Earth Observation (ThEO), Thünen Institute of Farm Economics, Braunschweig, Germany.\\
\textsuperscript{2} Earth Observation Lab, Geography Department, Humboldt-Universität zu Berlin, Unter den Linden 6, 10099 Berlin, Germany.\\
\textsuperscript{3} Earth and Life Institute, UCLouvain, 1348 Louvain-la-Neuve, Belgium.\\
\textsuperscript{4} Department of Agricultural Sciences, University of Helsinki, P.O. Box 28, FI-00014 Helsinki, Finland.\\
\textsuperscript{5} Center for Geospatial Analytics, North Carolina State University, NC 27695 Raleigh, USA.\\
\vspace{0.5em}
{\footnotesize
\textsuperscript{*}Correspondence: 
\href{mailto:jan.hemmerling@thuenen.de}{jan.hemmerling@thuenen.de}
}
\end{flushleft}
\vspace{1em}
\begin{abstract}

Organic farming is a key element in achieving more sustainable agriculture. For a better understanding of the development and impact of organic farming, comprehensive, spatially explicit information is needed. However, such data remain scarce in many countries. Despite the widespread use of remote sensing for mapping agricultural land use at the field level, its capability to distinguish organic and conventional systems has received little attention.
This study presents an approach for the discrimination of organic and conventional farming systems using intra-annual Sentinel-2 time series. In addition, it examines two factors influencing this discrimination: the joint learning of crop type information in a concurrent task and the role of spatial context. A Vision Transformer model based on the Temporo-Spatial Vision Transformer (TSViT) architecture was used to construct a classification model for the two farming systems. The model was extended for simultaneous learning of the crop type, creating a multitask learning setting. By varying the patch size presented to the model, we tested the influence of spatial context on the classification accuracy of both tasks.
We show that discrimination between organic and conventional farming systems using multispectral remote sensing data is feasible. However, classification performance varies substantially across crop types. For several crops, such as winter rye, winter wheat, and winter oat, F1 scores of 0.8 or higher can be achieved. In contrast, other agricultural land use classes, such as permanent grassland, orchards, grapevines, and hops, cannot be reliably distinguished, with F1 scores for the organic management class of 0.4 or lower. Joint learning of farming system and crop type provides only limited additional benefits over single-task learning. In contrast, incorporating wider spatial context improves the performance of both farming system and crop type classification. Overall, we demonstrate that a classification of agricultural farming systems is possible in a diverse agricultural region using multispectral remote sensing data.

\end{abstract}
\clearpage
\section{Introduction}

Organic farming is commonly regarded as a key instrument for achieving sustainability objectives in agriculture \citep{sanders_benefits_2025}. Benefits include reductions in greenhouse gas emissions, improvements in soil health, and enhanced biodiversity \citep{abbott_soil_2015,boone_environmental_2019, eyhorn_sustainability_2019, gamage_role_2023, mondelaers_meta-analysis_2009, tscharntke_beyond_2021}. Reflecting these advantages, the European Union has set the strategic goal of expanding organically managed land to 25\% of the total agricultural area by 2030 \citep{european_commission_european_2019}. These policy shifts prompt a critical examination of their potential environmental and socio-economic impacts. Accordingly, spatially explicit data on organic farming are essential for evaluating policy measures, tracking progress towards established targets, and modeling environmental effects \citep{houghton_carbon_2012, pongratz_models_2018}. Yet, spatially explicit information on the different farming systems (i.e., organic vs. conventional) remains sparse or incomplete, limiting robust assessments of the impacts associated with shifts in agricultural management.

Within the EU, organic farming is defined by regulations that prohibit practices such as the application of synthetic pesticides and fertilizers (Regulation (EU) 2018/848) \citep{european_parliament_and_council_of_the_european_union_regulation_2018}. As a result, organic systems must substitute synthetic inputs with mechanical and biological control measures, organic fertilization (e.g., compost, manure), and crop choices that support nutrient management and pest suppression \citep{barbieri_comparing_2017, chongtham2017factors, jalli2021effects, janicke2022field, kobierski2020effect, palaniappan2018organic}. Since several implicitly or explicitly regulated management components, such as the use of winter cover crops, tillage intensity, and fertilization, affect nutrient availability, crop phenology, and ultimately yield, their impacts can manifest in vegetation index trajectories and other spectral-temporal indicators observable in remote sensing data \citep{gao_mapping_2022, leo_combining_2023, lobert_unveiling_2025}. Consequently, remote sensing offers opportunities for filling the current knowledge gap regarding the spatial and temporal dynamics of organic agriculture \citep{ducati_application_2014, schuster_using_2023}.

A particularly promising data source here is the European Space Agency’s Sentinel-2 mission, which provides freely available multispectral imagery with high temporal, spatial, and spectral resolution. These data already enable detailed monitoring of crop type, phenology, canopy condition, and productivity at field scale \citep{hunt_high_2019, lobert_deep_2023, radeloff_need_2024, weiss_remote_2020, xie_retrieval_2019}, thereby offering the opportunity to evaluate whether management-induced differences could also be detected. However, it remains unclear whether organic and conventional systems yield spectral-temporal signatures that are sufficiently distinct and consistent for reliable discrimination. This challenge arises in part because farming system-specific practices, such as fertilization and pesticide applications, do not translate directly into unique spectral signatures. Fertilization responses, for example, are highly crop and growth stage specific and are strongly dependent on phenological timing, rendering them sensitive to the temporal resolution of satellite observations \citep{blaes_quantifying_2016, leo_combining_2023}. Pesticide-related signals may likewise be masked or confounded by environmental variability, including moisture conditions and vertical plant structure \citep{abdullah_present_2023}.

Initial studies, however, suggest that differences between organic and conventional management are expressed at multiple levels in satellite data, including crop spectral properties as well as their temporal and spatial patterns \citep{abdi_biodiversity_2021, atanasova_distinguishing_2021, denis_multispectral_2020}. In a comparative study of maize fields in Germany, Denis et al. (2020) showed using multispectral, hyperspectral, and in situ measurements that conventional fields exhibited higher chlorophyll and nitrogen content, greater plant height, and denser canopy cover, whereas organic fields were characterized by greater spatial heterogeneity. The authors concluded that management-induced plant traits underlie the observed spectral differences, indicating that variability in crop physiology translates into detectable signals in remote sensing imagery. Similarly, Atanasova et al. (2021) reported significant spectral differences between organically and conventionally managed wheat fields in Bulgaria based on Sentinel-2 data, with conventional fields exhibiting higher vegetation index values (e.g., NDVI) linked to biomass and yield differences \citep{ding_response_2022, shammi_use_2021}. Moreover, Zhou et al. (2025) demonstrated that management practices such as soil cover and tillage timing can be distinguished in optical and radar time series imagery, showing that management events themselves can leave detectable imprints in remote-sensing data \citep{zhou2025framework}. While these findings show that management-related differences can be reflected in spectral, temporal, and structural characteristics, they remain highly crop- and context-specific, and it is unclear whether such differences generalize across regions and crop types. Moreover, the presence of detectable practices or conditions does not uniquely indicate organic management, as conventionally managed fields may also employ more extensive or input-reducing practices, complicating the attribution of observed signals to a specific farming system \citep{lausch2025monitoring}. 

Methods for pixel-level remote sensing classifications have advanced from classical machine learning approaches, such as Random Forests and Support Vector Machines, to deep learning architectures, including U-Net, 3D CNNs, and Vision Transformers \citep{han_survey_2023, li_review_2024}. Remote sensing has adapted methodologies from related disciplines, such as video processing, where time series techniques are frequently adapted \citep{feng_water_2018, wei_multi-temporal_2019, zhu_deep_2017}. In video, where variation is typically greater across space than over time, data are usually decomposed into spatial and temporal components, where spatial context is processed first and temporal dynamics are added thereafter \citep{jiao_transformer_2023, khan_transformers_2022, yuan_sits-former_2022}. This is reflected in architectures such as U-TAE, which combine convolutional spatial encoding with attention-based temporal aggregation and are inspired by spatial–temporal factorization strategies commonly used in video processing \citep{fare_garnot_panoptic_2021}.

Remote sensing differs, as temporal variability at fixed spatial locations often plays a more prominent role than spatial variability. As a result, class-defining properties are primarily captured in spectral-temporal dynamics rather than spatial patterns, necessitating adaptations when applying video-based methods to remote sensing data \citep{rolf_mission_2024}. While pixel-based time series classification captures spectral and temporal dynamics \citep{blickensdorfer_mapping_2022}, incorporating spatial context provides additional neighborhood information that can disambiguate spectrally and temporally similar classes and has been shown to improve overall classification accuracy \citep{lu_ai_2024, pereira_chessmix_2021}. The Temporo-Spatial Vision Transformer (TSViT) introduced by \citet{tarasiou2023vits} offers an architecture explicitly designed to model temporal dynamics together with spatial context in satellite image time series. The model first prioritizes spectral-temporal representations and then situates them within their spatial surroundings. Building on the findings of Denis et al. (2020), TSViT is well suited for integrating spatial information into both crop type and farming system classification tasks.
Both crop type and farming system jointly characterize agricultural landscapes, with each influencing how the other is expressed in spectral and temporal remote sensing signals. The spectral-temporal signatures of crop type and management practice are expected to be difficult to disentangle; however, some management methods may be crop-specific. Multitask learning, understood as the simultaneous optimization of related prediction tasks within a single model, enables the extraction of crop-specific spectral-temporal patterns while also capturing management-related features that generalize across crops. Prior work has shown that auxiliary tasks can guide models toward more robust feature learning \citep{fang_mscpunet_2024, waldner_deep_2020}. By training the model to classify crop type and farming system concurrently, such synergistic effects can be evaluated regarding their potential to improve the discrimination of farming systems and crop types.

Although evidence suggests that management-related differences may be reflected in spectral-temporal remote sensing data, there is limited understanding of how crop type and spatial context jointly influence the differentiation of farming systems. Closing these gaps is essential for advancing more holistic remote sensing–based assessments of agricultural farming systems. In this paper, we therefore address the following research questions: (i) can the farming system (organic vs. conventional) be distinguished from intra-annual multispectral Sentinel-2 time series, and how does separability vary across crop types? (ii) does the joint learning of crop types alongside the farming system (multitask) improve performance relative to singletask models? (iii) what is the influence of spatial context on the accuracy of classifying crop type and farming system?

\section{Data and Methods}

\subsection{Study area and reference data}

\begin{figure}[t]
\centering

\includegraphics[scale=0.99]{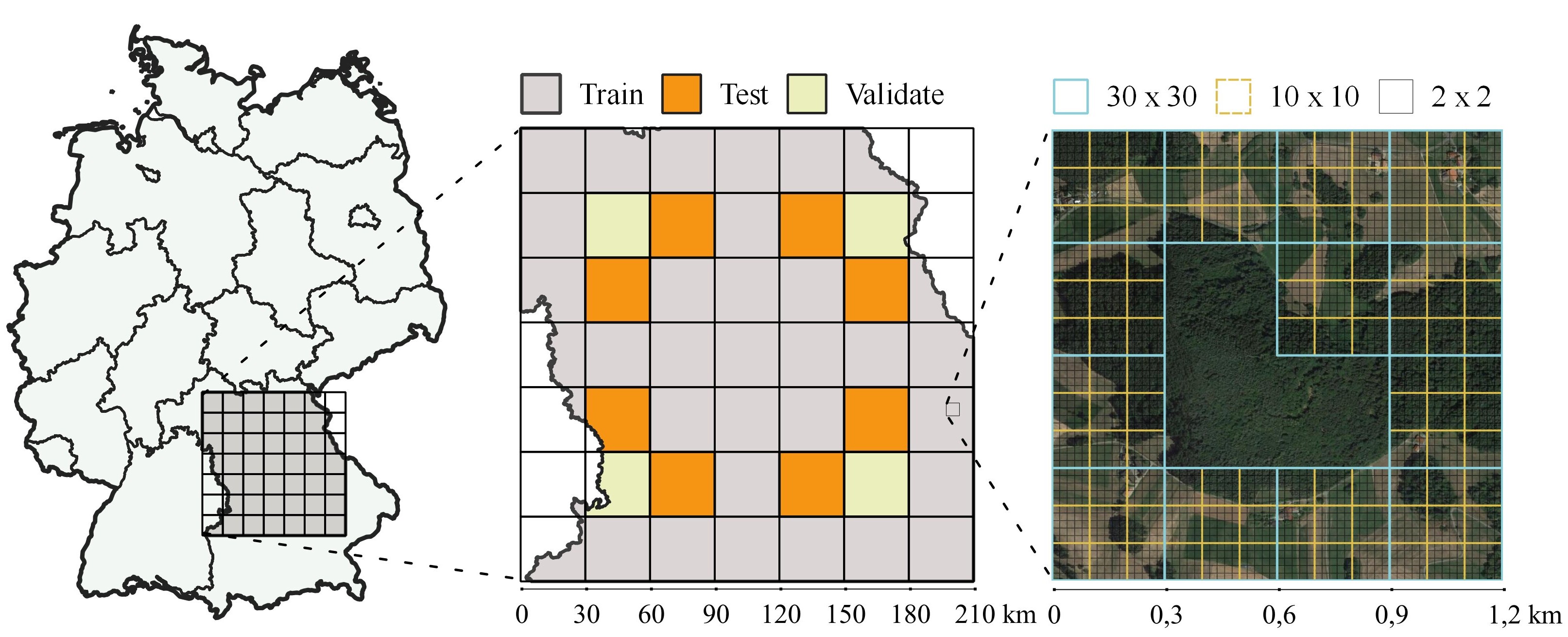}
\caption{Location of our study area in Germany, Zoom-in (1): grid system with 30 × 30 km tiles, indicated are training, validation and test tiles; Zoom-in (2): Subdivision of tiles into 300 × 300 m patches (30 × 30 pixel); outlined are patches with more than 15\% agricultural land as used for training and validation, further subdivision of the patches into 100 × 100 m (10 × 10 pixel) and 20× 20 m (2 × 2 pixel), Background: Google Earth}
\label{fig:study_area}
\end{figure}

The study region is located in the south-eastern German federal state of Bavaria and covers 38,521 km², comparable in size to Switzerland \textit{(Figure~\ref{fig:study_area})}. It comprises nearly one-fifth of Germany’s agricultural area (approximately 3.09 million hectares) of which about 66\% is arable land \citep{bayerisches_staatsministerium_fur_ernahrung_landwirtschaft_forsten_und_tourismus_stmelf_bayerischer_2024}. Bavaria extends from relatively warm and dry lowland regions in the north to cool, high-precipitation Alpine and pre-Alpine zones in the south, encompassing flat glacial end-moraine areas and the Danube lowlands in the south, the low mountain ranges of the Bavarian and Bohemian Forests in the east, and the Franconian Forest in the north. As a result, Bavaria is characterized by pronounced climatic and edaphic gradients, together with marked contrasts in soil types, topography, and growing conditions \citep{dhillon_landscape_2025}.

For this region, reference data of agricultural land use were retrieved from the Integrated Administration and Control System (IACS), developed by the European Union for managing agricultural subsidy programs \citep{toth_spatial_2016}. The dataset provides parcel-level land-use information with more than 270 distinct crop-type labels for the reference year 2022. We aggregated these into 23 crop-type categories, including permanent grassland, largely following the scheme proposed by \cite{blickensdorfer_mapping_2022}. In addition to crop type, we differentiated two farming systems, conventional and organic farming. Organic management follows the legally defined EU regulations \citep{EU_Organic_2007} and is explicitly indicated in the IACS data due to separate subsidy provisions. According to this data, 47.6\% of the study area is used for agricultural purposes, of which 9.24\% was under organic management. Permanent grassland (22.9\%), maize (19.1\%), winter wheat (19.1\%), and winter barley (9.5\%) account for the largest shares of agricultural area in the study region \textit{(Table 1)}. Crops with the highest proportions under organic management include broad bean (74.0\%), lupin (65.5\%), and spring oat (46.9\%), though each occupies less than 1\% of the total area. Additional crops with relatively high organic shares are sunflower (30.9\%), soy (29.3\%), and cultivated grassland (28.7\%).

\begin{table}[ht]
 \caption{Total area, share of agricultural area and area under organic management per crop type in the study area}
\centering
\begin{tabular}{lrrr}
\\[-2ex]
 \textbf{Crop\:type} & $Area [ha]$ & $Share [\%]$ & $Org, [\%]$ \\ 
  \hline & \\[-2ex]
Winter wheat & 34336.24 & 19.05 & 7.98 \\ 
Winter barley & 17152.44 & 9.52 & 1.71 \\ 
Winter rye & 3442.15 & 1.91 & 18.48 \\ 
Triticale & 6453.90 & 3.58 & 9.35 \\ 
Spring barley & 6110.98 & 3.39 & 11.70 \\ 
Spring oat & 1785.68 & 0.99 & 46.92 \\ 
Maize & 34478.31 & 19.13 & 2.74 \\ 
Potato & 3269.97 & 1.81 & 5.26 \\ 
Sugar beet & 5028.13 & 2.79 & 3.92 \\ 
Winter rapeseed & 6295.24 & 3.49 & 0.45 \\ 
Sunflower & 368.75 & 0.20 & 30.86 \\ 
Cultivated grassland & 9029.91 & 5.01 & 28.73 \\ 
Permanent grassland & 41203.14 & 22.86 & 11.46 \\ 
Vegetables & 1300.02 & 0.72 & 20.76 \\ 
Peas & 1085.08 & 0.60 & 26.67 \\ 
Broad bean & 425.72 & 0.24 & 74.02 \\ 
Lupin & 71.61 & 0.04 & 65.49 \\ 
Soy & 1006.26 & 0.56 & 29.34 \\ 
Hops & 1753.84 & 0.97 & 0.97 \\ 
Grapevine & 422.03 & 0.23 & 12.02 \\ 
Other agricultural areas & 924.90 & 0.51 & 23.07 \\ 
Orchard & 558.66 & 0.31 & 18.46 \\ 
Fallow land & 3720.97 & 2.06 & 11.91 \\ 
   \hline
\end{tabular}
\end{table}

\begin{table}[ht]
   \caption{Data Distribution of Data Sets} 
   \label{tab:example}
   \small
   \centering
   \begin{tabular}{lcr}
   \toprule
   \textbf{} & \textbf{Tiles [n]} & \textbf{Patches [n]}  \\ 
   \midrule
         Training Data & 37 & 218815 \\
         Validation Data & 4 & 26079 \\
         Test Data & 8 & 80000\\
    \bottomrule
   \end{tabular}
\end{table}
The study area was divided into a regular grid of 30 × 30 km, yielding 49 tiles. Of these, 37 were used for training, 4 for validation, and 8 for testing \textit{(Figure~\ref{fig:study_area})}. Test and validation tiles were regularly distributed outside the outer boundary regions of the study area in order to improve the representativeness of validation and test samples. Each tile was further subdivided into 300 × 300 m patches containing the full 2020 time series, which were stored as intermediate inputs for model training \textit{(Table 2)}.

We evaluated the influence of spatial context on both farming system and crop type classification. We chose a patch size of 300 × 300 m as a compromise between computational feasibility and adequate field coverage, given that 50\% of the fields have a maximum field span of about 200 m in our study area. To assess the effect of spatial context, we also conducted additional experiments with 100 × 100 m and 20 × 20 m patches. For training and validation, we used only 300 × 300 m patches containing more than 15\% agricultural area according to the reference data, whereas no such restriction was applied to the test set.

\subsection{Sentinel-2 data}
We acquired all available Sentinel-2 scenes with less than 75\% cloud coverage over the study area between November 2019 and February 2021. Image pre-processing was conducted using the Framework for Operational Radiometric Correction for Environmental Monitoring (FORCE) \citep{frantz_forcelandsat_2019}, including radiometric and geometric correction, co-registration, and cloud and shadow masking with the improved Fmask algorithm \citep{frantz_improvement_2018, rufin_operational_2020, zhu_improvement_2015}. The 20 m Sentinel-2 bands were sharpened to 10 m spatial resolution using the spectral-only setting of the ImproPhe algorithm \citep{frantz_improving_2016}.

We interpolated Sentinel-2 observations to an equidistant 10-day time series using an RBF filter ensemble with five kernels($\sigma \in {5, 10, 16, 32, 64}$) \citep{schwieder_mapping_2016}. Kernel weights were adapted to observation density, with double weighting applied to the smallest kernels to better preserve abrupt changes, such as tillage or harvest \citep{blickensdorfer_mapping_2022}. Scenes from November–December 2019 and January–February 2021 were included to reduce data gaps at the start and end of the time series, and only the interpolated images for 2020 were retained. The resulting dataset comprises 10 spectral bands at 36 time steps with 10 m spatial resolution.

\subsection{Model}

\begin{figure}[t]
\centering
\includegraphics[width=\textwidth]{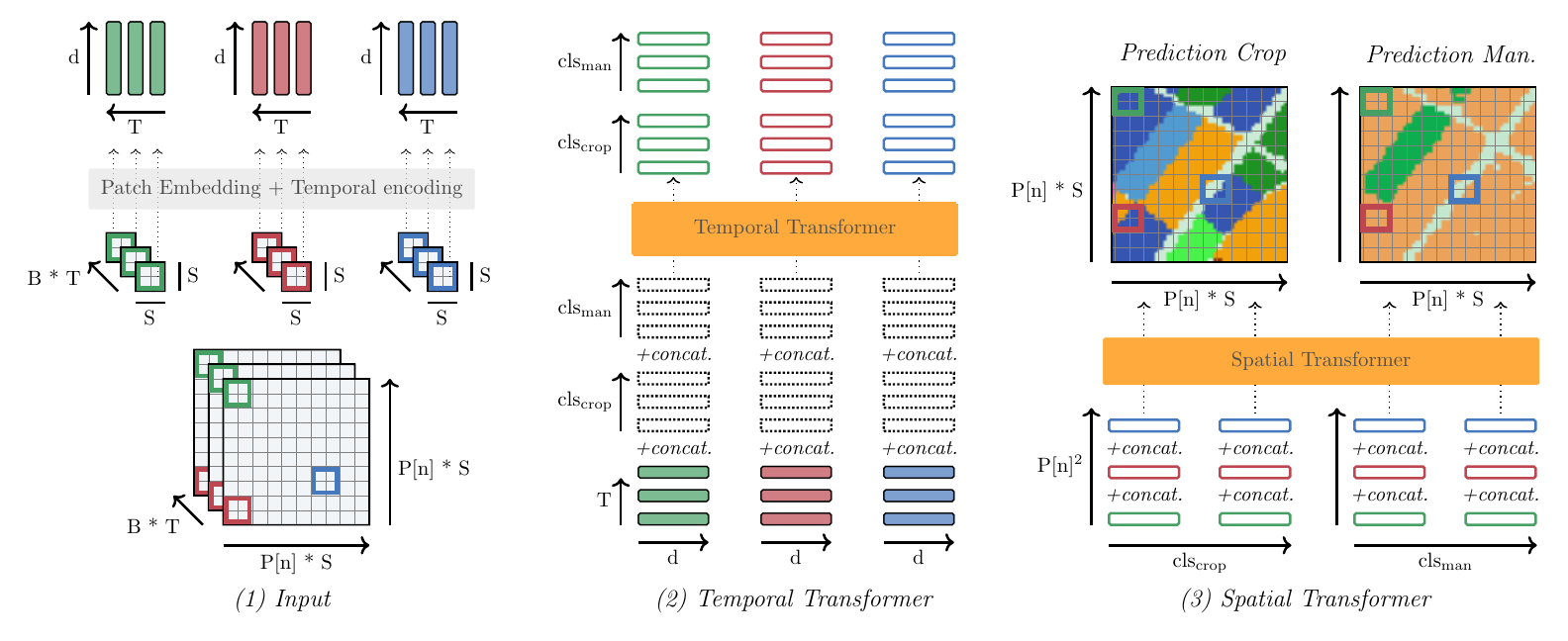}
\caption{Overview of the adapted Temporo-Spatial Vision Transformer (TSViT), Pixel are represented by grey grid lines, colored boxes represent sub-patches (1) Input patches are arranged into n sub-patches (PP) of size S. For each of the 36 time steps (T) 10 bands (B) are embedded into tokens of length d (2) In the temporal transformer stage, tokens interact across T and the concatenated class tokens (3) In the spatial transformer stage, class tokens interact only with their corresponding spatial class token. Final predictions for crop type and management system are produced via separate argmax layers.}
\label{fig:tsvit_architecture}
\end{figure}

The model used in this study builds on the Temporo-Spatial Vision Transformer (TSViT) introduced by \cite{tarasiou2023vits}. Below, we summarize the core architecture, describe our modifications, and specify the model training settings. For a detailed exposition of the original TSViT architecture, we refer the reader to \cite{tarasiou2023vits}.
TSViT consists of two consecutive encoder stages that process the input along the temporal and spatial dimensions using stacks of multi-layer transformer blocks \textit{(Figure~\ref{fig:tsvit_architecture})}. As in other transformer architectures, TSViT processes the data as a sequence of tokens, meaning that each acquisition is converted into a vector embedding on which the model performs attention operations. Each input patch is divided into non-overlapping sub-patches (P), each containing the full time series (T, here 36 time steps with 10 spectral bands, B). Each acquisition is embedded into vectors of length d, forming sequences of length T. For each of the n classes, a learnable class token of length d is appended, yielding representations of size (T + n) × d. These sequences are processed by the temporal encoder, which models dependencies across time while allowing interaction between time series tokens and class tokens \textit{(Figure 2.2)}. After this stage, only the class tokens are retained, while the time series tokens are discarded. In the subsequent spatial encoder, class tokens from all sub-patches are processed jointly, enabling spatial interaction and defining the model’s global receptive field \textit{(Figure 2.3)}. The resulting tokens are projected to unnormalized class scores.
For multitask adaptation, we concatenate two distinct sets of class tokens before the temporal encoder: one for crop type classification (23 classes) and one for management classification (3 classes). In the spatial stage, class tokens interact only within their respective class, following the original TSViT restriction that prevents cross-class token mixing. After the spatial encoder, tokens are grouped by task to yield two independent sets of unnormalized class scores. Training is carried out using separate cross-entropy losses for crop type and management, which are aggregated into a global loss.

\subsection{Training settings}
\begin{figure}[ht]
\centering
\resizebox{0.800\textwidth}{!}{\includegraphics{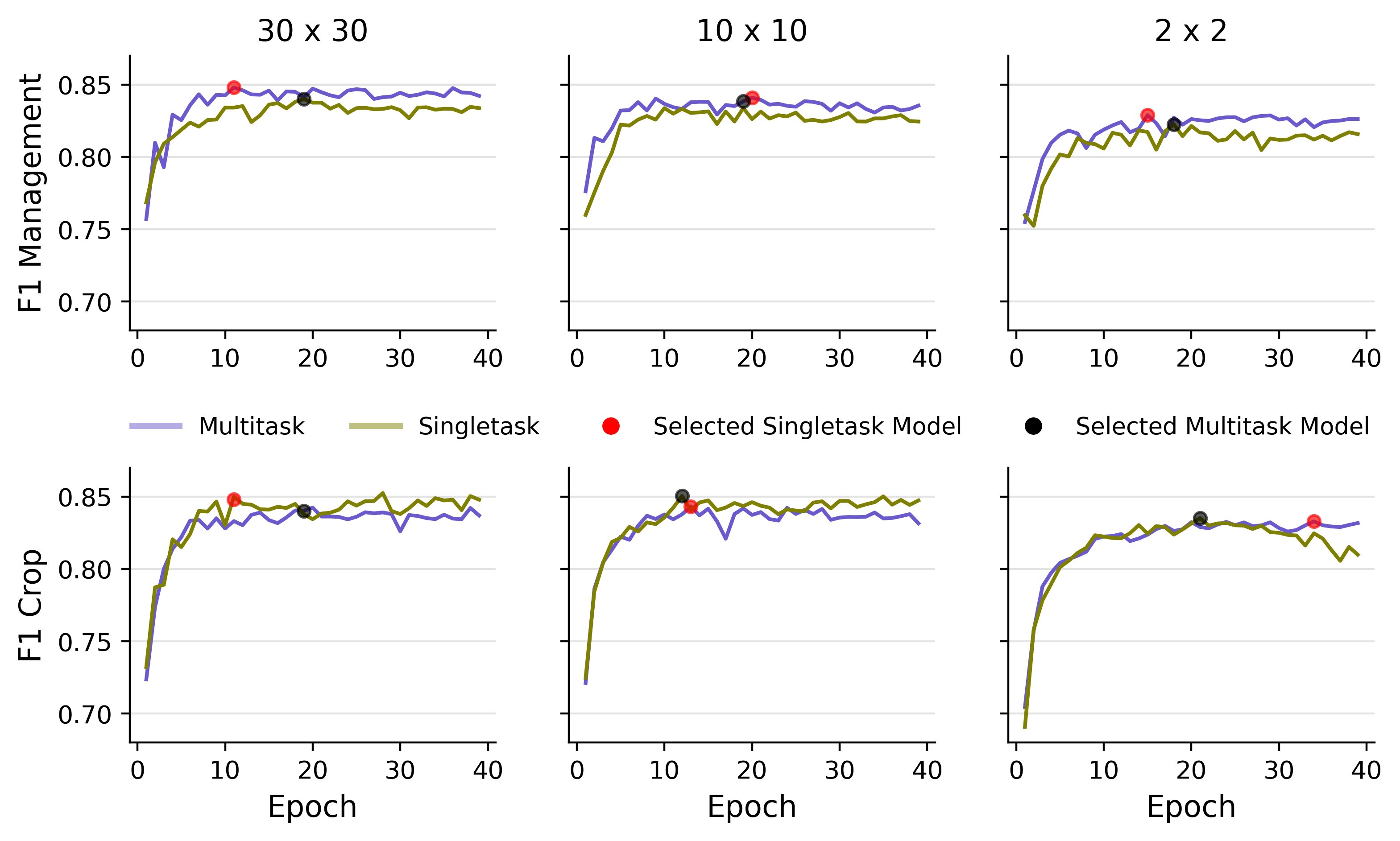}}
\caption{Mean F1-score derived from validation data set during model training after each epoch.}
\label{fig:training_curve}
\end{figure}

Following \cite{tarasiou2023vits}, we used 2 × 2 pixel sub-patches, which were converted into tokens of length 150. The temporal encoder stage consisted of eight transformer blocks, and the spatial encoder stage of four blocks, each with eight attention heads. For experiments with a total patch size of 2 × 2 pixels, the spatial stage was omitted. All experiments were conducted with a fixed batch size of 20, corresponding to 20 patches of 30 × 30 pixels (300 × 300 m). For smaller spatial extents, the batch size was kept constant by subdividing each sample into non-overlapping 10 × 10 or 2 × 2 pixel patches, yielding 9 (100 × 100 m) or 15 (20 × 20 m) sub-patches per patch, respectively. This ensured that each batch contained the same total amount of spectral and temporal information across experiments. To reduce the over-representation of the Background class, only patches with more than 15\% agricultural area were used for training and validation. In contrast, the test set included all patches without filtering, resulting in a higher share of the Background class.
Data were normalized using the 5\% and 95\% quantiles, derived from 1000 randomly sampled patches. Model training employed the AdamW optimizer with a cosine learning rate schedule: the learning rate increased linearly from $5 \times 10^{-5}$ to $1 \times 10^{-4}$ over 2 warm-up epochs, followed by a cosine decay to $1 \times 10^{-5}$ by epoch 20 \citep{loshchilov_decoupled_2017, loshchilov_sgdr_2016}. All models were trained for 40 epochs with identical settings across runs to ensure comparability. Mixed precision was used to accelerate training without affecting performance \citep{dettmers_gpt3_2022}. All experiments were conducted on an NVIDIA RTX A6000 GPU.
Models for further evaluation were selected based on the best validation F1-scores recorded after each epoch \textit{(Figure~\ref{fig:training_curve})}. For comparisons between multitask and singletask models, multitask models were chosen according to the task-specific F1-score: the best management mean F1-score for farming system classification and the best crop type mean F1-score for crop type mapping.

\subsection{Random Forest Baseline}
As a baseline, we trained a Random Forest \citep{breiman_random_2001} model in addition to the TSViT models, as it represents one of the most established classifiers for large-area crop type classification \citep{blickensdorfer_mapping_2022, ghassemi_designing_2022}. Unlike the patch-based TSViT, the Random Forest is trained on single pixels. For each field in the training area one random sample was drawn, applying a 15 m inward buffer to each field polygon to avoid boundary effects \citep{blickensdorfer_mapping_2022, griffiths2019intra}. For each 30 × 30 km tile, 500 additional samples were drawn from the Background class (see Appendix A2 and A3 for sample distribution).

\subsection{Accuracy Measures}
All models were evaluated on the spatially independent test tiles. For each class, F1-scores, precision, and recall were calculated \citep{foody_status_2002}. Overall performance was assessed using the macro-averaged F1-score, which balances class-specific accuracies equally, i.e. irrespective of class frequency, and overall accuracy, which reflects general map quality.

\section{Results}
We compared overall accuracy, mean, and class-wise F1-scores for crop type and farming system classification using predictions from the trained TSViT models. The analysis differentiated between input patch sizes of 30 × 30, 10 × 10, and 2 × 2 pixels. For each spatial extent, we contrasted multitask models with singletask models for both classification tasks. In total, 12 TSViT models (six per task) were evaluated and compared against a Random Forest baseline.

\begin{table*}[htb]
\footnotesize
\centering
\newcolumntype{Z}{>{\small}c}
\tabcolsep=0.15cm 
\caption{Classification results, F1-scores, Recall and Precision values for single crop types }
\resizebox{\textwidth}{!}{ %
\begin{tabular}{@{}p{2.35cm}ccc|ccc|ccc|ccc|ccc|ccc|ccc@{}}
\toprule
\textbf{} & \multicolumn{6}{c|}{\textbf{30 x 30}} & \multicolumn{6}{c|}{\textbf{10 x 10}} & \multicolumn{6}{c|}{\textbf{2 x 2}} & \multicolumn{3}{c}{\textbf{1 x 1}} \\
\midrule
\textbf{} & \multicolumn{3}{c|}{\textbf{Multitask}} & \multicolumn{3}{c|}{\textbf{Singletask}} & \multicolumn{3}{c|}{\textbf{Multitask}} & \multicolumn{3}{c|}{\textbf{Singletask}} & \multicolumn{3}{c|}{\textbf{Multitask}} & \multicolumn{3}{c|}{\textbf{Singletask}} & \multicolumn{3}{c}{\textbf{RF}} \\
\midrule
\textbf{Crop Types} & F1 & Re. & Pr . & F1 & Re. & Pr.  & F1 & Re. & Pr.  & F1 & Re. & Pr.  & F1 & Re. & Pr.  & F1 & Re. & Pr.  & F1 & Re. & Pr. \\
\midrule
Winter wheat & 0.94 & 0.94 & 0.95 & \textbf{0.95} & 0.94 & 0.95 & 0.94 & 0.94 & 0.94 & 0.94 & 0.94 & 0.94 & 0.93 & 0.93 & 0.94 & 0.92 & 0.92 & 0.93 & 0.80 & 0.72 & 0.90\\
Winter barley & 0.95 & 0.95 & 0.96 & \textbf{0.96} & 0.95 & 0.96 & 0.95 & 0.95 & 0.95 & 0.95 & 0.95 & 0.95 & 0.95 & 0.94 & 0.95 & 0.94 & 0.93 & 0.94 &0.79& 0.75& 0.84  \\
Winter rye & 0.83 & 0.86 & 0.81 & \textbf{0.84} & 0.85 & 0.83 & 0.81 & 0.81 & 0.81 & 0.82 & 0.84 & 0.80 & 0.80 & 0.84 & 0.77 & 0.78 & 0.79 & 0.76 & 0.20 & 0.40 & 0.13\\
Triticale & 0.83 & 0.82 & 0.83 & \textbf{0.84} & 0.83 & 0.86 & 0.81 & 0.82 & 0.81 & 0.81 & 0.83 & 0.79 & 0.79 & 0.80 & 0.79 & 0.77 & 0.79 & 0.75 & 0.40 & 0.58 & 0.31 \\

Spring oat & \textbf{0.76} & 0.75 & 0.77 & \textbf{0.76} & 0.73 & 0.78 & 0.75 & 0.76 & 0.74 & \textbf{0.76} & 0.76 & 0.76 & 0.72 & 0.75 & 0.70 & 0.70 & 0.65 & 0.76 & 0.33 & 0.49 & 0.25 \\
Maize & \textbf{0.97} & 0.96 & 0.97 & \textbf{0.97} & 0.96 & 0.98 & \textbf{0.97} & 0.96 & 0.97 & \textbf{0.97} & 0.97 & 0.97 & \textbf{0.97} & 0.96 & 0.97 & 0.96 & 0.96 & 0.97 & 0.90 & 0.86 & 0.93\\

Sugar beet & \textbf{0.97} & 0.97 & 0.97 & \textbf{0.97} & 0.97 & 0.98 & \textbf{0.97} & 0.96 & 0.97 & \textbf{0.97} & 0.96 & 0.97 & 0.96 & 0.96 & 0.97 & 0.96 & 0.96 & 0.96 & 0.91 & 0.93 & 0.89\\
Winter rapeseed & \textbf{0.97} & 0.97 & 0.97 & \textbf{0.97} & 0.97 & 0.98 & \textbf{0.97} & 0.97 & 0.97 & \textbf{0.97} & 0.97 & 0.97 & \textbf{0.97} & 0.97 & 0.97 & 0.96 & 0.96 & 0.97 & 0.91 & 0.89 & 0.92\\
Sunflower & 0.81 & 0.79 & 0.83 & 0.86 & 0.86 & 0.85 & 0.82 & 0.88 & 0.77 & 0.84 & 0.88 & 0.81 & 0.79 & 0.80 & 0.79 & 0.78 & 0.80 & 0.75 & 0.36 & 0.80 & 0.23 \\
Cult. grassl. & \textbf{0.76} & 0.75 & 0.76 & \textbf{0.76} & 0.77 & 0.76 & 0.75 & 0.74 & 0.76 & \textbf{0.76} & 0.76 & 0.76 & 0.74 & 0.77 & 0.71 & 0.72 & 0.76 & 0.69 & 0.48 & 0.60 & 0.40\\
Perm. grassl. & 0.85 & 0.81 & 0.89 & 0.85 & 0.81 & 0.89 & \textbf{0.86} & 0.86 & 0.85 & \textbf{0.86} & 0.83 & 0.89 & \textbf{0.86} & 0.84 & 0.88 & 0.85 & 0.82 & 0.88 & 0.67 & 0.51 & 0.95 \\
Vegetables & \textbf{0.82} & 0.84 & 0.80 & \textbf{0.82} & 0.80 & 0.83 & 0.80 & 0.76 & 0.83 & 0.81 & 0.82 & 0.80 & 0.79 & 0.79 & 0.79 & 0.76 & 0.76 & 0.77 & 0.41 & 0.35 & 0.49\\

Broad bean & \textbf{0.86} & 0.92 & 0.81 & \textbf{0.86} & 0.88 & 0.84 & 0.85 & 0.86 & 0.84 & \textbf{0.86} & 0.90 & 0.82 & 0.84 & 0.84 & 0.84 & 0.83 & 0.87 & 0.79 & 0.61 & 0.89 & 0.46 \\

Soy & 0.89 & 0.92 & 0.86 & \textbf{0.91} & 0.95 & 0.88 & 0.89 & 0.92 & 0.85 & 0.90 & 0.93 & 0.87 & 0.88 & 0.94 & 0.82 & 0.87 & 0.92 & 0.83 & 0.63 & 0.86 & 0.49 \\
Hops & \textbf{0.97} & 0.97 & 0.97 & \textbf{0.97} & 0.96 & 0.98 & 0.96 & 0.96 & 0.96 & \textbf{0.97} & 0.97 & 0.96 & 0.96 & 0.97 & 0.96 & 0.96 & 0.97 & 0.96 & 0.78 & 0.99 & 0.65 \\
Grapevine & \textbf{0.75} & 0.67 & 0.85 & \textbf{0.75} & 0.68 & 0.82 & 0.71 & 0.61 & 0.86 & 0.74 & 0.66 & 0.84 & 0.74 & 0.69 & 0.80 & 0.67 & 0.63 & 0.72 & 0.13 & 0.07 & 0.69\\

Orchard & \textbf{0.34} & 0.49 & 0.26 & 0.33 & 0.52 & 0.24 & 0.31 & 0.49 & 0.23 & 0.32 & 0.53 & 0.23 & 0.27 & 0.59 & 0.17 & 0.26 & 0.57 & 0.17 & 0.02 & 0.08 & 0.01 \\

Background & \textbf{0.96} & 0.97 & 0.95 & \textbf{0.96} & 0.97 & 0.95 & \textbf{0.96} & 0.96 & 0.96 & \textbf{0.96} & 0.97 & 0.96 & \textbf{0.96} & 0.96 & 0.96 & \textbf{0.96} & 0.96 & 0.95 & 0.88 & 0.99 & 0.80\\
\midrule\midrule
Mean F1 & 0.82 & 0.84 & 0.81 & \textbf{0.83} & 0.84 & 0.82 & 0.81 & 0.83 & 0.81 & 0.82 & 0.84 & 0.81 & 0.81 & 0.83 & 0.79 & 0.79 & 0.81 & 0.78 & 0.54 & 0.66 & 0.54 \\
\toprule
Overall accuracy & \textbf{0.93} &  & & \textbf{0.93} &  &  & \textbf{0.93} &  &  & \textbf{0.93} &  &  & \textbf{0.93} &  &  & 0.92 &  &  & 0.80 &  &  \\
\toprule
\end{tabular}}
\end{table*}

All TSViT-based models achieved an overall accuracy of 93\%, except the 2 × 2 singletask model (92\%) \textit{(Table 3)}. Mean F1-scores only varied slightly within and between classification tasks and patch sizes: the best result was obtained with the 30 × 30 singletask model (0.83), followed by the 30 × 30 multitask and 10 × 10 singletask models (0.82). The 2 × 2 singletask model showed the lowest performance (0.79). The Random Forest baseline performed considerably worse (OA = 80\%, mean F1 = 0.54).
Across most crop types, performances of singletask models were slightly better than multitask configurations for 30 × 30 and 10 × 10 inputs. For 2 × 2 patches, however, multitask models performed better in 21 of 24 classes. Seven crop types (Winter wheat, Winter barley, Maize, Potato, Sugar beet, Winter rapeseed, and Hops) consistently achieved F1-scores above 0.9 across all TSViT models, demonstrating robust separability. In contrast, the Orchards class consistently yielded the lowest performance (F1 < 0.5). The Background class was reliably separated from agricultural classes across all TSViT models (F1 = 0.96).

\begin{table*}[htb]
\footnotesize

\centering
\newcolumntype{Z}{>{\small}c}
\tabcolsep=0.15cm 
\caption{Accuracy measures for management system classification }
\resizebox{\textwidth}{!}{ 
\begin{tabular}{@{}p{2.35cm}ccc|ccc|ccc|ccc|ccc|ccc|ccc@{}}
\toprule
\textbf{} & \multicolumn{6}{c|}{\textbf{30 x 30}} & \multicolumn{6}{c|}{\textbf{10 x 10}} & \multicolumn{6}{c|}{\textbf{2 x 2}} & \multicolumn{3}{c}{\textbf{1 x 1}} \\
\midrule
\textbf{} & \multicolumn{3}{c|}{\textbf{Multitask}} & \multicolumn{3}{c|}{\textbf{Singletask}} & \multicolumn{3}{c|}{\textbf{Multitask}} & \multicolumn{3}{c|}{\textbf{Singletask}} & \multicolumn{3}{c|}{\textbf{Multitask}} & \multicolumn{3}{c|}{\textbf{Singletask}} & \multicolumn{3}{c}{\textbf{RF}} \\
\midrule
\textbf{Management} & F1 & Re. & Pr. & F1 & Re. & Pr. & F1 & Re. & Pr. & F1 & Re. & Pr. & F1 & Re. & Pr. & F1 & Re. & Pr. & F1 & Re. & Pr. \\
\midrule
Conventional & \textbf{0.93} & 0.91 & 0.95 & \textbf{0.93} & 0.91 & 0.95 & \textbf{0.93} & 0.92 & 0.95 & \textbf{0.93} & 0.92 & 0.94 & \textbf{0.93} & 0.91 & 0.95 & 0.92 & 0.91 & 0.94 & 0.79 & 0.93 & 0.69 \\
Organic & \textbf{0.61} & 0.69 & 0.55 & 0.59 & 0.64 & 0.54 & 0.59 & 0.67 & 0.52 & 0.58 & 0.65 & 0.52 & 0.55 & 0.70 & 0.46 & 0.53 & 0.63 & 0.45 & 0.25 & 0.15 & 0.74 \\
Background & \textbf{0.96} & 0.97 & 0.95 & \textbf{0.96} & 0.97 & 0.95 & \textbf{0.96} & 0.97 & 0.96 & \textbf{0.96} & 0.97 & 0.96 & \textbf{0.96} & 0.97 & 0.95 & \textbf{0.96} & 0.96 & 0.95 & 0.92 & 0.96 & 0.89 \\
\midrule\midrule
Mean F1 & 0.83 & 0.86 & 0.81 & 0.82 & 0.84 & 0.81 & 0.83 & 0.85 & 0.81 & 0.82 & 0.85 & 0.81 & 0.81 & 0.86 & 0.79 & 0.80 & 0.83 & 0.78 & 0.65 & 0.68 & 0.77 \\
\toprule
Overall accuracy &  0.93 &  &  & 0.93 &  & & 0.94 &  &  & 0.94 &  &  & 0.93 &  &  & 0.93 &  &  &  0.80 &  &  \\
\toprule
\end{tabular}
}
\end{table*}
For management system classification, the highest overall accuracies (94\%) were achieved by the 10 $\times$ 10 multitask and singletask models \textit{(Table 4)}. All other TSViT model runs reached 93\% overall accuracy, whereas the Random Forest baseline achieved  80\%. Mean F1-scores were highest for the 30 × 30 and 10 $\times$ 10 multitask models (0.83), with multitask models showing a slight but consistent advantage ($\approx$ +0.01) over singletask models. The Background and Conventional class remained stable across models (F1 $\approx$ 0.96 and 0.93, respectively). The Organic class showed dependence on spatial context, with F1-scores increasing from 0.53 (2 × 2 singletask) to 0.61 (30 × 30 multitask).

\begin{figure}[!t]
\centering
\includegraphics[width=\textwidth]{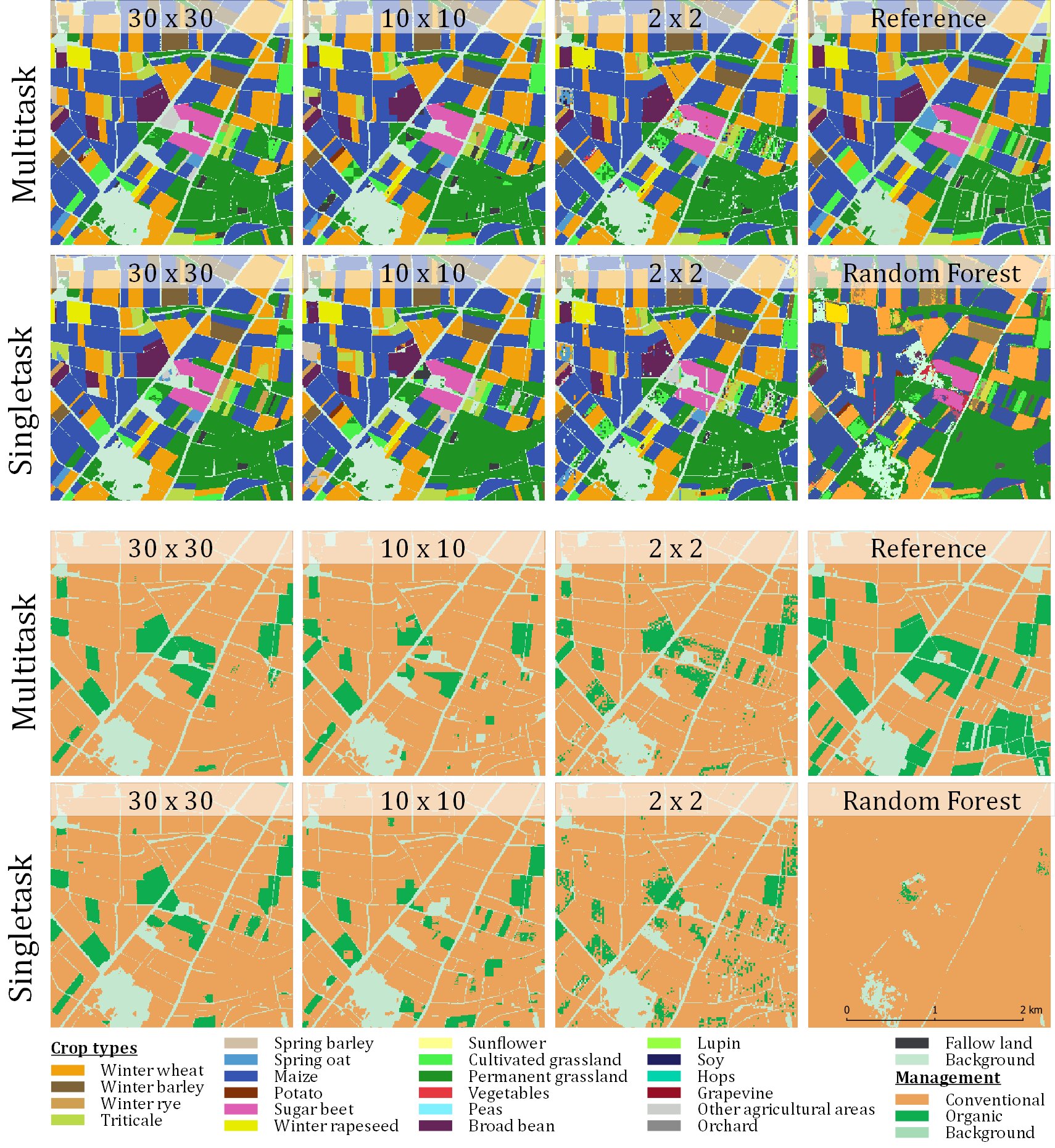}
\caption{Crop type and management system predictions from model runs with different spatial input dimensions (30 x 30, 10 x 10, 2 x 2 pixel, as described in 2.3, IACS data as reference, Random Forest model predictions as described in 2.8)}
\label{fig:results_maps}
\end{figure}

Figure 4 illustrates example classification outputs for crop type (top) and farming system (bottom), comparing TSViT multitask and singletask models across the three spatial input dimensions (30 × 30, 10 × 10, and 2 × 2 pixels), alongside the IACS reference and the Random Forest baseline. The TSViT models show high visual agreement with the reference data, particularly for the 30 × 30 and 10 × 10 inputs, where field boundaries and within-field structures are preserved. In contrast, the 2 × 2 models exhibit a more pronounced salt-and-pepper effect and spatially incoherent pixel-level misclassifications in both crop type and farming system classification. Visually, the improvement from 2 × 2 to 10 × 10 appears larger than the improvement from 10 × 10 to 30 × 30, which is consistent with the observed changes in the F1-scores \textit{(Table 4)}. For the farming system classification task, the 2 × 2 multitask model shows fewer small-scale misclassifications than its singletask counterpart; aside from this difference, multitask and singletask outputs are generally comparable. The Random Forest baseline largely aligns with the reference for crop types but produces less coherent field structures; for management form, it markedly overestimates the conventional class and frequently misclassifies agricultural areas as Background.

\begin{figure}[htbp]
\centering
\includegraphics[width=\textwidth]{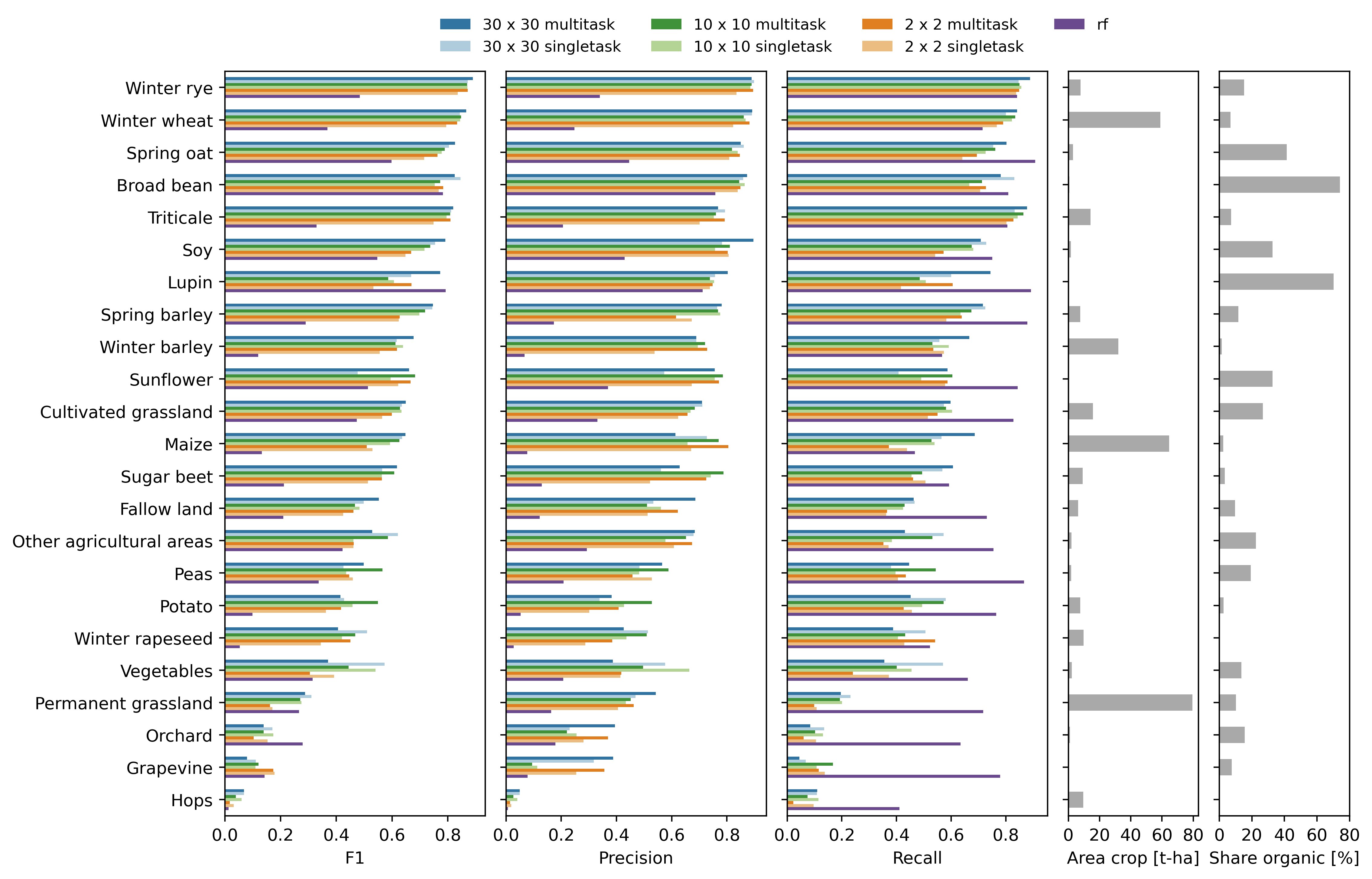}
\caption{Crop type and management form predictions from model runs with different spatial input dimensions (30 x 30, 10 x 10, 2 x 2 pixel, as described in 2.3, IACS data as reference, Random Forest model predictions as described in 2.8)}
\label{fig:results}
\end{figure}

The accuracy of the organic management classification performance varied substantially across crop types \textit{(Figure~\ref{fig:results})}. For five crops (winter rye, winter wheat, spring oat, broad bean, and triticale) F1-scores above 0.8 were achieved across all TSViT models, while hops, grapevine, and orchards consistently remained below 0.2. lupine and vegetables showed the greatest variability between model types, with lupine performing best in the 30 × 30 multitask setting (F1 = 0.58, compared to 0.46 in the 2 × 2 singletask), and vegetables consistently favoring singletask models. Random Forest achieved higher F1-scores than TSViT for orchards (0.02–0.34 vs. 0.40) and lupine (0.46–0.58 vs. 0.61), but generally suffered from low precision (0.45–0.50), despite at times attaining higher recall values.

\section{Discussion}
In this study, we investigated the potential of TSViT models for classifying crop species and distinguishing between conventional and organic farming based on Sentinel-2 time series, and evaluated the effects of (a) multitask learning and (b) spatial input extent. To address (a), we compared singletask and multitask versions of the model. To address (b), we trained TSViT models with different spatial extents (30 × 30, 10 × 10, and 2 × 2 pixels), while keeping the spectral-temporal profile constant. All results were compared with a Random Forest baseline.

\subsection*{Farming system classification}
To our knowledge, this is the first study to systematically assess the capability of multispectral satellite time series to classify agricultural farming systems (organic vs. conventional) using a deep learning framework. Our results show that while farming system classification is feasible, performance varies strongly across crop type classes. For crops such as winter rye, winter wheat, broad bean, spring oat, and triticale, F1-scores above 0.8 demonstrate that management-induced differences are sufficiently pronounced to be captured by remote sensing based models. In contrast, especially for perennials like hops, grapevine, and orchards, classification accuracy remained low, suggesting that management effects are subtle. The underlying reasons for the large variation in classification accuracy across crop types remain largely unclear. We assume that, for some crops, management practices such as fertilization, pesticide application, or crop density produce more pronounced and thus more detectable differences in growth dynamics, whereas for others these effects are more subtle. This appears particularly relevant for perennial crops like grapevine and hops. Perennial systems typically retain high biomass throughout the growing season, and the dominant spatial field patterns in vineyards and hop yards tend to remain structurally consistent across farming systems. Moreover, not all potentially relevant covariates, such as site-specific factors including soil type, farm size, or fine-scale topography, were explicitly accounted for in this study, which may further contribute to the observed variability.

Our results can also be interpreted in light of crop-specific yield gaps between organic and conventional farming \citep{de_ponti_crop_2012}. De Ponti et al. (2012) showed that cereals and pulses generally exhibit larger yield differences (organic yields 70–80\% of conventional), while permanent crops display greater variability in yield reductions, reflecting the more heterogeneous responses of perennial systems to management. This pattern supports our findings: cereals such as winter wheat, rye, and triticale, as well as legumes like broad bean, showed the highest farming system classification accuracies (F1 > 0.8), whereas grapevine and orchards remained poorly separable (F1 < 0.2). It is likely that the same management factors driving yield gaps, such as nutrient stress and differences in pest and disease pressure, also shape spectral-temporal signals detectable by remote sensing. This interpretation is supported by \cite{abdi_biodiversity_2021}, who demonstrated that management-induced physiological differences can produce detectable spectral variability. Consequently, crops with larger and more consistent yield gaps between organic and conventional management tend to be more distinguishable in classification tasks, whereas those with weaker or more variable yield contrasts are less reliably separable.

The occurrence of organic farming varied strongly across crop types, with broad bean and lupine exceeding 70\% share of organic farming, while winter barley, maize, and hops remained below a share of 3\% of the total agricultural area \textit{(Table 1,Figure~\ref{fig:results})}. Training, validation, and test data generally reflected this distribution, though discrepancies existed (e.g., vegetables: 32.7\% organic farming in validation vs. 13.6\% in test). Nevertheless, performance patterns cannot be explained by the share of farming systems alone, as both strong and weak results occurred in classes with high representation of organic farming. Instead, classes with larger absolute extents of organically managed area typically benefited from greater sample availability, even when their relative shares were low (e.g., maize, with a small share of organic farming areas but still substantially larger total organic farming area than other crops such as lupine with a high share of organic farming). This suggests that crop-specific sample availability, in addition to class imbalance, is a driver for model performance.

\subsection*{Impacts of multitask learning}
Multitask learning did not improve overall classification accuracy. While multitask models yielded slightly higher F1-scores for farming system classification, they consistently underperformed in crop type classification. However, multitask models converged faster \textit{(Figure~\ref{fig:training_curve})}, which may stem from the higher number of learnable tokens introduced in the multitask setting (one set of class tokens for crop type and an additional set for management form). Additional learnable tokens can simplify optimization by providing explicit aggregation points and stable representation anchors, which may lead to faster convergence during training \citep{darcet_vision_2023, dosovitskiy_image_2020}. These results suggest that crop type and management form are largely separable tasks and can be learned independently. Whether multitask models primarily learn crop-specific differences in management or general management traits across crops remains unclear and warrants further study.

\subsection*{Influence of spatial context}
Our results confirm that the spatial input extent affects classification performance. Larger spatial inputs (30 × 30 pixels) improved mean F1-scores for both crop type and management form classification. For instance, the crop type singletask model improved by 0.04 in mean F1-score and the management singletask model by 0.02 when moving from 2 × 2 to 30 × 30 inputs. These improvements are consistent with earlier findings that contextual information mitigates local misclassifications, similar to post-processing approaches based on majority votes \citep{turkoglu_crop_2021}. Although these gains are relatively modest, the comparison with the spatially agnostic Random Forest baseline shows markedly stronger performance for the TSViT models, again emphasizing the role of spatial context.

However, accuracy benefits of spatial context were not uniform across classes. Crop types such as Lupine, Triticale, and Sunflower benefited most from an increased spatial extent, whereas Permanent Grassland and the Background class showed little to no improvement. Notably, the step from 2 × 2 to 10 × 10 patches produced the largest gains, while improvements from 10 × 10 to 30 × 30 were comparatively modest, suggesting a threshold beyond which additional context adds limited value in this landscape. It is worth noting that the 2 × 2 setting also differs architecturally, as the spatial transformer stage is omitted and the overall model size is smaller, which may contribute to the observed jump in performance. Contextual effects are particularly visible at field edges, where mixed spectral signals can still be correctly classified if sufficient spatial context is available. However, excessive context may also introduce errors by overriding locally correct predictions, as seen for Permanent Grassland, which performs worse at 30 × 30 pixel inputs. These edge-related effects are not uniform across crops and highlight a trade-off between capturing broader field structure and preserving fine-scale spectral detail.

In farming system classification, the impact of spatial context was less apparent in overall accuracy but more pronounced in class-specific results, especially for the Organic class. Larger patch sizes improved precision, reducing false positives, while recall tended to decline. This trade-off is likely driven by class imbalance, where the model minimizes losses by favoring the majority class (Conventional), thus improving precision but lowering recall for Organic. Moreover, organic-specific signals may be strongest at fine spatial scales but become diluted when aggregated with surrounding context. These patterns suggest that management practices leave both fine-scale (e.g., canopy vigor, weed cover) and field-scale (e.g., crop vigor, structural characteristics) signatures, with their detectability depending on patch size. Crop-specific dependencies were also evident: while winter rye, winter wheat, and triticale were largely insensitive to patch size, classes such as soy, cultivated grassland, and maize showed clear improvements with larger spatial contexts. This indicates that management practices (e.g., fertilizer or pesticide omission) influence the spatial–spectral signature of crops in ways that extend beyond simple spatial smoothing effects.

The Random Forest baseline achieved an overall accuracy of 80\%, comparable to similar studies using pixel-based classifiers at national or continental scales \citep{blickensdorfer_mapping_2022, ghassemi_designing_2022}. Given the smaller study area in our case, higher accuracies might have been expected. We attribute the discrepancy to study design choices, particularly the relatively high number of crop type classes, especially the inclusion of a background class and the use of a wall-to-wall validation approach, which can be challenging for pixel-based methods. Class-wise performance patterns also matched earlier reports, with strong variability across crops (e.g., low F1-scores for sunflower). However, the TSViT models delivered substantial performance gains over Random Forest across nearly all crop types, with more balanced precision and recall behavior. Notably, TSViT achieved a remarkable improvement for cultivated grassland, where F1-scores increased substantially compared to Random Forest predictions.

The Random Forest’s relatively high precision for the Organic class (0.74) indicates the presence of spectral-temporal signals associated with management, consistent with \cite{atanasova_distinguishing_2021} who linked phenological changes to fertilization practices. However, TSViT models outperformed Random Forest across all evaluated settings, achieving simultaneously higher precision and recall as well as higher overall accuracy and mean F1. In contrast, Random Forest often overpredicted minority classes, resulting in inflated recall but substantially lower precision.

In summary, our findings demonstrate that spatial context improves crop type and farming system classification, particularly when increasing input size from very small (2 × 2) to moderate (10 × 10) patches. Gains diminish at larger scales, and in some cases excessive context may even reduce accuracy. Farming system classification benefits from larger spatial context primarily through improved precision for the Organic class. The TSViT model consistently outperformed Random Forest, especially in balancing recall and precision across classes. Multitask learning offered limited benefits but faster convergence, indicating that architectural choices rather than shared representations may drive most performance gains in this setting. Moreover, by capturing long-range dependencies, Transformers can exploit field-scale variability, often higher in organic systems and reduced in conventional systems due to synthetic inputs, thus making management-related characteristics more learnable beyond local pixel neighborhoods.

\subsection*{Limitations}
Crop composition, management intensity, and yield levels vary substantially across regions in Germany, and it remains uncertain to what extent the crop-specific separability observed in Bavaria holds under different environmental and agronomic conditions. In particular, soil type and land-use intensity were not considered as explanatory factors, which may cause low-input conventional and organic systems to exhibit overlapping spectral-temporal characteristics. Extending the analysis to multi-year and multi-regional datasets will therefore be essential for assessing robustness. This may be particularly relevant for organic systems, which are characterized by distinct and often longer-term crop rotations \citep{barbieri_comparing_2017}, suggesting that classification performance may improve when multi-year temporal context is taken into account.

Methodologically, several limitations should be acknowledged. The use of equidistant time series may obscure short-term phenological signals, potentially limiting discrimination between farming systems; future work should explore non-equidistant or phenology-aware sampling schemes. Furthermore, although we examined spatial contexts up to 30 × 30 pixels, the potential benefits of larger input sizes remain untested due to the approx. quadratic increase in computational cost in the spatial transformer stage. The relatively small spatial extent of our study further constrains geographical generalizability and underscores the need for larger-scale evaluations, ideally at national scale, to capture regional variation in crop composition, field structure, and management practices.

Finally, the limited advantages observed for multitask learning suggest that crop type and farming system signals can be learned largely independently within the present dataset and model configuration. How this finding transfers to other regions, sensor types, or modeling strategies remains an open question.

\section{Conclusion}
To our knowledge, this study is the first to demonstrate that agricultural farming systems (organic vs. conventional) can be classified from Sentinel-2 multispectral time series using deep learning. The results show that management-related signals are detectable and that robust classification is achievable for several major crop groups, particularly cereals and legumes, which exhibit high separability. These crops therefore hold strong potential for operational monitoring of organic farming adoption. At the same time, classification performance remains strongly crop-dependent. Permanent and fruit crops, such as orchards, grapevine, hops, and permanent grassland, proved more challenging, indicating that management effects are less pronounced or less consistently expressed in their spectral-temporal patterns. 

Overall, our findings confirm earlier indications that distinguishing organic and conventional management with remote sensing is feasible and extend this understanding by providing a crop-resolved assessment. Future work should incorporate multi-year observations and crop rotation information to further strengthen classification performance and enhance generalizability across regions and production systems.

\section*{Acknowledgement}
This publication was prepared within the framework of the nationwide monitoring program of biodiversity in agricultural landscapes (MonViA), commissioned and funded by the German Federal Ministry of Agriculture, Food and Regional Identity (BMELH). It was further supported by the German Ministry for Economic Affairs and Climate Action through the EOekoLand Project under Grant 50RP2230A. The authors would like to thank Dr. Judith Brüggemann from the Research Institute of Organic Agriculture (FiBL) for the valuable discussions on the design of the dataset and the compilation of the reference data.

\bibliographystyle{plainnat}
\bibliography{references}

\begin{thebibliography}{73}
\providecommand{\natexlab}[1]{#1}
\providecommand{\url}[1]{\texttt{#1}}
\expandafter\ifx\csname urlstyle\endcsname\relax
  \providecommand{\doi}[1]{doi: #1}\else
  \providecommand{\doi}{doi: \begingroup \urlstyle{rm}\Url}\fi

\bibitem[Abbott and Manning(2015)]{abbott_soil_2015}
Lynette~K Abbott and David~{AC} Manning.
\newblock Soil health and related ecosystem services in organic agriculture.
\newblock 4\penalty0 (3), 2015.

\bibitem[Abdi et~al.(2021)Abdi, Carrié, Sidemo-Holm, Cai, Boke-Olén, Smith,
  Eklundh, and Ekroos]{abdi_biodiversity_2021}
Abdulhakim~M. Abdi, Romain Carrié, William Sidemo-Holm, Zhanzhang Cai, Niklas
  Boke-Olén, Henrik~G. Smith, Lars Eklundh, and Johan Ekroos.
\newblock Biodiversity decline with increasing crop productivity in
  agricultural fields revealed by satellite remote sensing.
\newblock 130:\penalty0 108098, 2021.
\newblock ISSN 1470-160X.
\newblock \doi{10.1016/j.ecolind.2021.108098}.
\newblock URL
  \url{https://www.sciencedirect.com/science/article/pii/S1470160X21007639}.

\bibitem[Abdullah et~al.(2023)Abdullah, Mohana, Khan, Ahmed, Hossain, Islam,
  Redoy, Ferdush, Bhuiyan, and Hossain]{abdullah_present_2023}
Hasan~M. Abdullah, Nusrat~T. Mohana, Bhoktear~M. Khan, Syed~M. Ahmed, Maruf
  Hossain, {KH}~Shakibul Islam, Mahadi~H. Redoy, Jannatul Ferdush, {MAHB}
  Bhuiyan, and Motaher~M. Hossain.
\newblock Present and future scopes and challenges of plant pest and disease
  (p\&d) monitoring: Remote sensing, image processing, and artificial
  intelligence perspectives.
\newblock 32:\penalty0 100996, 2023.
\newblock {ISBN}: 2352-9385.

\bibitem[Atanasova et~al.(2021)Atanasova, Bozhanova, Biserkov, and
  Maneva]{atanasova_distinguishing_2021}
Dina Atanasova, Violeta Bozhanova, Valko Biserkov, and Vasilina Maneva.
\newblock Distinguishing areas of organic, biodynamic and conventional farming
  by means of multispectral images. a pilot study.
\newblock 35\penalty0 (1):\penalty0 977--993, 2021.

\bibitem[Barbieri et~al.(2017)Barbieri, Pellerin, and
  Nesme]{barbieri_comparing_2017}
Pietro Barbieri, Sylvain Pellerin, and Thomas Nesme.
\newblock Comparing crop rotations between organic and conventional farming.
\newblock 7, 2017.
\newblock URL \url{https://api.semanticscholar.org/CorpusID:30440395}.

\bibitem[{Bayerisches Staatsministerium für Ernährung, Landwirtschaft,
  Forsten und Tourismus
  (StMELF)}(2024)]{bayerisches_staatsministerium_fur_ernahrung_landwirtschaft_forsten_und_tourismus_stmelf_bayerischer_2024}
{Bayerisches Staatsministerium für Ernährung, Landwirtschaft, Forsten und
  Tourismus (StMELF)}.
\newblock Bayerischer agrarbericht 2024, 2024.
\newblock URL
  \url{https://www.stmelf.bayern.de/mam/cms01/agrarpolitik/dateien/stmelf_aktuell100_agrarbericht.pdf}.

\bibitem[Blaes et~al.(2016)Blaes, Chomé, Lambert, Traoré, Schut, and
  Defourny]{blaes_quantifying_2016}
Xavier Blaes, Guillaume Chomé, Marie-Julie Lambert, Pierre~Sibiry Traoré,
  Antonius~{GT} Schut, and Pierre Defourny.
\newblock Quantifying fertilizer application response variability with {VHR}
  satellite {NDVI} time series in a rainfed smallholder cropping system of
  mali.
\newblock 8\penalty0 (6):\penalty0 531, 2016.
\newblock {ISBN}: 2072-4292.

\bibitem[Blickensdörfer et~al.(2022)Blickensdörfer, Schwieder, Pflugmacher,
  Nendel, Erasmi, and Hostert]{blickensdorfer_mapping_2022}
Lukas Blickensdörfer, Marcel Schwieder, Dirk Pflugmacher, Claas Nendel, Stefan
  Erasmi, and Patrick Hostert.
\newblock Mapping of crop types and crop sequences with combined time series of
  sentinel-1, sentinel-2 and landsat 8 data for germany.
\newblock 269:\penalty0 112831, 2022.
\newblock ISSN 0034-4257.
\newblock \doi{10.1016/j.rse.2021.112831}.
\newblock URL
  \url{https://www.sciencedirect.com/science/article/pii/S0034425721005514}.

\bibitem[Boone et~al.(2019)Boone, Roldán-Ruiz, Muylle, Dewulf, and
  {others}]{boone_environmental_2019}
Lieselot Boone, Isabel Roldán-Ruiz, Hilde Muylle, Jo~Dewulf, and {others}.
\newblock Environmental sustainability of conventional and organic farming:
  Accounting for ecosystem services in life cycle assessment.
\newblock 695:\penalty0 133841, 2019.

\bibitem[Breiman(2001)]{breiman_random_2001}
Leo Breiman.
\newblock Random forests.
\newblock 45:\penalty0 5--32, 2001.

\bibitem[Chongtham et~al.(2017)Chongtham, Bergkvist, Watson, Sandstr{\"o}m,
  Bengtsson, and {\"O}born]{chongtham2017factors}
Iman~Raj Chongtham, G{\"o}ran Bergkvist, Christine~A Watson, Emil
  Sandstr{\"o}m, Jan Bengtsson, and Ingrid {\"O}born.
\newblock Factors influencing crop rotation strategies on organic farms with
  different time periods since conversion to organic production.
\newblock \emph{Biological Agriculture \& Horticulture}, 33\penalty0
  (1):\penalty0 14--27, 2017.

\bibitem[{Council of the European Union}(2007)]{EU_Organic_2007}
{Council of the European Union}.
\newblock Council regulation (ec) no 834/2007 of 28 june 2007 on organic
  production and labelling of organic products and repealing regulation (eec)
  no 2092/91.
\newblock Official Journal of the European Union, L 189, 20.7.2007, pp. 1--23,
  2007.
\newblock URL
  \url{https://eur-lex.europa.eu/legal-content/EN/TXT/?uri=CELEX:32007R0834}.

\bibitem[Darcet et~al.(2023)Darcet, Oquab, Mairal, and
  Bojanowski]{darcet_vision_2023}
Timothée Darcet, Maxime Oquab, Julien Mairal, and Piotr Bojanowski.
\newblock Vision transformers need registers.
\newblock 2023.

\bibitem[De~Ponti et~al.(2012)De~Ponti, Rijk, and
  Van~Ittersum]{de_ponti_crop_2012}
Tomek De~Ponti, Bert Rijk, and Martin~K. Van~Ittersum.
\newblock The crop yield gap between organic and conventional agriculture.
\newblock 108:\penalty0 1--9, 2012.
\newblock {ISBN}: 0308-521X.

\bibitem[Denis et~al.(2020)Denis, Desclee, Migdall, Hansen, Bach, Ott, Kouadio,
  and Tychon]{denis_multispectral_2020}
Antoine Denis, Baudouin Desclee, Silke Migdall, Herbert Hansen, Heike Bach,
  Pierre Ott, Amani~Louis Kouadio, and Bernard Tychon.
\newblock Multispectral remote sensing as a tool to support organic crop
  certification: Assessment of the discrimination level between organic and
  conventional maize.
\newblock 13\penalty0 (1):\penalty0 117, 2020.

\bibitem[Dettmers et~al.(2022)Dettmers, Lewis, Belkada, and
  Zettlemoyer]{dettmers_gpt3_2022}
Tim Dettmers, Mike Lewis, Younes Belkada, and Luke Zettlemoyer.
\newblock Gpt3. int8 (): 8-bit matrix multiplication for transformers at scale.
\newblock 35:\penalty0 30318--30332, 2022.

\bibitem[Dhillon et~al.(2025)Dhillon, Koellner, Asam, Bogenreuther, Dech,
  Gessner, Gruschwitz, Annuth, Kraus, and Rummler]{dhillon_landscape_2025}
Maninder~Singh Dhillon, Thomas Koellner, Sarah Asam, Jakob Bogenreuther, Stefan
  Dech, Ursula Gessner, Daniel Gruschwitz, Sylvia~Helena Annuth, Tanja Kraus,
  and Thomas Rummler.
\newblock Landscape structure, climate variability, and soil quality shape crop
  biomass patterns in agricultural ecosystems of bavaria.
\newblock 16:\penalty0 1630087, 2025.
\newblock {ISBN}: 1664-462X.

\bibitem[Ding et~al.(2022)Ding, He, Zhou, Hu, Cai, Wang, Li, Xu, and
  Shi]{ding_response_2022}
Yibo Ding, Xiaofeng He, Zhaoqiang Zhou, Jie Hu, Huanjie Cai, Xiaoyun Wang,
  Lusheng Li, Jiatun Xu, and Haiyun Shi.
\newblock Response of vegetation to drought and yield monitoring based on
  {NDVI} and {SIF}.
\newblock 219:\penalty0 106328, 2022.
\newblock {ISBN}: 0341-8162.

\bibitem[Dosovitskiy et~al.(2020)Dosovitskiy, Beyer, Kolesnikov, Weissenborn,
  Zhai, Unterthiner, Dehghani, Minderer, Heigold, Gelly, and
  {others}]{dosovitskiy_image_2020}
Alexey Dosovitskiy, Lucas Beyer, Alexander Kolesnikov, Dirk Weissenborn,
  Xiaohua Zhai, Thomas Unterthiner, Mostafa Dehghani, Matthias Minderer, Georg
  Heigold, Sylvain Gelly, and {others}.
\newblock An image is worth 16x16 words: Transformers for image recognition at
  scale.
\newblock 2020.

\bibitem[Ducati et~al.(2014)Ducati, Sarate, and
  Fachel]{ducati_application_2014}
Jorge~R. Ducati, Rafael~E. Sarate, and Jandyra~{MG} Fachel.
\newblock Application of remote sensing techniques to discriminate between
  conventional and organic vineyards in the loire valley, france.
\newblock 48\penalty0 (3):\penalty0 135--144, 2014.
\newblock {ISBN}: 2494-1271.

\bibitem[{European Commission}(2019)]{european_commission_european_2019}
{European Commission}.
\newblock The european green deal, 2019.
\newblock URL \url{https://eur-lex.europa.eu/eli/COM/2019/640/oj}.

\bibitem[{European Parliament and Council of the European
  Union}(2018)]{european_parliament_and_council_of_the_european_union_regulation_2018}
{European Parliament and Council of the European Union}.
\newblock Regulation ({EU}) 2018/848 of the european parliament and of the
  council of 30 may 2018 on organic production and labelling of organic
  products and repealing council regulation ({EC}) no 834/2007, 2018.
\newblock URL \url{https://eur-lex.europa.eu/eli/reg/2018/848/oj}.
\newblock Series: {OJ} L 150.

\bibitem[Eyhorn et~al.(2019)Eyhorn, Muller, Reganold, Frison, Herren,
  Luttikholt, Mueller, Sanders, Scialabba, Seufert, and
  {others}]{eyhorn_sustainability_2019}
Frank Eyhorn, Adrian Muller, John~P Reganold, Emile Frison, Hans~R Herren,
  Louise Luttikholt, Alexander Mueller, Jürn Sanders, Nadia El-Hage Scialabba,
  Verena Seufert, and {others}.
\newblock Sustainability in global agriculture driven by organic farming.
\newblock 2\penalty0 (4):\penalty0 253--255, 2019.

\bibitem[Fang et~al.(2024)Fang, Zhang, Han, Yang, Luo, Liu, Zhang, and
  Wang]{fang_mscpunet_2024}
Kedi Fang, Shengwei Zhang, Yongting Han, Lin Yang, Meng Luo, Lu~Liu, Qian
  Zhang, and Bo~Wang.
\newblock {MSCPUnet}: A multi-task neural network for plot-level crop
  classification in complex agricultural areas.
\newblock 9:\penalty0 100660, 2024.
\newblock {ISBN}: 2772-3755.

\bibitem[Fare~Garnot and Landrieu(2021)]{fare_garnot_panoptic_2021}
Vivien~Sainte Fare~Garnot and Loic Landrieu.
\newblock Panoptic segmentation of satellite image time series with
  convolutional temporal attention networks.
\newblock In \emph{2021 {IEEE}/{CVF} International Conference on Computer
  Vision ({ICCV})}, pages 4852--4861. {IEEE}, 2021.
\newblock ISBN 978-1-6654-2812-5.
\newblock \doi{10.1109/ICCV48922.2021.00483}.
\newblock URL \url{https://ieeexplore.ieee.org/document/9711189/}.

\bibitem[Feng et~al.(2018)Feng, Sui, Huang, Xu, and An]{feng_water_2018}
Wenqing Feng, Haigang Sui, Weiming Huang, Chuan Xu, and Kaiqiang An.
\newblock Water body extraction from very high-resolution remote sensing
  imagery using deep u-net and a superpixel-based conditional random field
  model.
\newblock 16\penalty0 (4):\penalty0 618--622, 2018.

\bibitem[Foody(2002)]{foody_status_2002}
Giles~M. Foody.
\newblock Status of land cover classification accuracy assessment.
\newblock 80\penalty0 (1):\penalty0 185--201, 2002.
\newblock {ISBN}: 0034-4257.

\bibitem[Frantz(2019)]{frantz_forcelandsat_2019}
David Frantz.
\newblock {FORCE}—landsat+ sentinel-2 analysis ready data and beyond.
\newblock 11\penalty0 (9):\penalty0 1124, 2019.

\bibitem[Frantz et~al.(2016)Frantz, Stellmes, Röder, Udelhoven, Mader, and
  Hill]{frantz_improving_2016}
David Frantz, Marion Stellmes, Achim Röder, Thomas Udelhoven, Sebastian Mader,
  and Joachim Hill.
\newblock Improving the spatial resolution of land surface phenology by fusing
  medium-and coarse-resolution inputs.
\newblock 54\penalty0 (7):\penalty0 4153--4164, 2016.
\newblock {ISBN}: 0196-2892.

\bibitem[Frantz et~al.(2018)Frantz, Haß, Uhl, Stoffels, and
  Hill]{frantz_improvement_2018}
David Frantz, Erik Haß, Andreas Uhl, Johannes Stoffels, and Joachim Hill.
\newblock Improvement of the fmask algorithm for sentinel-2 images: Separating
  clouds from bright surfaces based on parallax effects.
\newblock 215:\penalty0 471--481, 2018.

\bibitem[Gamage et~al.(2023)Gamage, Gangahagedara, Gamage, Jayasinghe,
  Kodikara, Suraweera, and Merah]{gamage_role_2023}
Ashoka Gamage, Ruchira Gangahagedara, Jeewan Gamage, Nepali Jayasinghe,
  Nathasha Kodikara, Piumali Suraweera, and Othmane Merah.
\newblock Role of organic farming for achieving sustainability in agriculture.
\newblock 1\penalty0 (1):\penalty0 100005, 2023.
\newblock {ISBN}: 2949-9119.

\bibitem[Gao et~al.(2022)Gao, Zhang, Yun, Ji, Ma, Wang, Li, and
  Zhu]{gao_mapping_2022}
Lulu Gao, Chao Zhang, Wenju Yun, Wenjun Ji, Jiani Ma, Huan Wang, Cheng Li, and
  Dehai Zhu.
\newblock Mapping crop residue cover using adjust normalized difference residue
  index based on sentinel-2 {MSI} data.
\newblock 220:\penalty0 105374, 2022.
\newblock {ISBN}: 0167-1987.

\bibitem[Ghassemi et~al.(2022)Ghassemi, Dujakovic, Żółtak, Immitzer,
  Atzberger, and Vuolo]{ghassemi_designing_2022}
Babak Ghassemi, Aleksandar Dujakovic, Mateusz Żółtak, Markus Immitzer,
  Clement Atzberger, and Francesco Vuolo.
\newblock Designing a european-wide crop type mapping approach based on machine
  learning algorithms using {LUCAS} field survey and sentinel-2 data.
\newblock 14\penalty0 (3):\penalty0 541, 2022.

\bibitem[Griffiths et~al.(2019)Griffiths, Nendel, and
  Hostert]{griffiths2019intra}
Patrick Griffiths, Claas Nendel, and Patrick Hostert.
\newblock Intra-annual reflectance composites from sentinel-2 and landsat for
  national-scale crop and land cover mapping.
\newblock \emph{Remote sensing of environment}, 220:\penalty0 135--151, 2019.

\bibitem[Han et~al.(2023)Han, Zhang, Wang, Wang, Huang, Li, Wang, Chen, Li,
  Feng, and {others}]{han_survey_2023}
Wei Han, Xiaohan Zhang, Yi~Wang, Lizhe Wang, Xiaohui Huang, Jun Li, Sheng Wang,
  Weitao Chen, Xianju Li, Ruyi Feng, and {others}.
\newblock A survey of machine learning and deep learning in remote sensing of
  geological environment: Challenges, advances, and opportunities.
\newblock 202:\penalty0 87--113, 2023.

\bibitem[Houghton et~al.(2012)Houghton, House, Pongratz, Van Der~Werf, Defries,
  Hansen, Le~Quéré, and Ramankutty]{houghton_carbon_2012}
Richard~A Houghton, Jo~I House, Julia Pongratz, Guido~R Van Der~Werf, Ruth~S
  Defries, Matthew~C Hansen, Corinne Le~Quéré, and Navin Ramankutty.
\newblock Carbon emissions from land use and land-cover change.
\newblock 9\penalty0 (12):\penalty0 5125--5142, 2012.

\bibitem[Hunt et~al.(2019)Hunt, Blackburn, Carrasco, Redhead, and
  Rowland]{hunt_high_2019}
Merryn~L Hunt, George~Alan Blackburn, Luis Carrasco, John~W Redhead, and
  Clare~S Rowland.
\newblock High resolution wheat yield mapping using sentinel-2.
\newblock 233:\penalty0 111410, 2019.

\bibitem[Jalli et~al.(2021)Jalli, Huusela, Jalli, Kauppi, Niemi, Himanen, and
  Jauhiainen]{jalli2021effects}
Marja Jalli, Erja Huusela, Heikki Jalli, Katja Kauppi, Mari Niemi, Sari
  Himanen, and Lauri Jauhiainen.
\newblock Effects of crop rotation on spring wheat yield and pest occurrence in
  different tillage systems: a multi-year experiment in finnish growing
  conditions.
\newblock \emph{Frontiers in Sustainable Food Systems}, 5:\penalty0 647335,
  2021.

\bibitem[J{\"a}nicke et~al.(2022)J{\"a}nicke, Goddard, Stein, Steinmann, Lakes,
  Nendel, and M{\"u}ller]{janicke2022field}
Clemens J{\"a}nicke, Adam Goddard, Susanne Stein, Horst-Henning Steinmann,
  Tobia Lakes, Claas Nendel, and Daniel M{\"u}ller.
\newblock Field-level land-use data reveal heterogeneous crop sequences with
  distinct regional differences in germany.
\newblock \emph{European Journal of Agronomy}, 141:\penalty0 126632, 2022.

\bibitem[Jiao et~al.(2023)Jiao, Zhang, Liu, Liu, Yang, Ma, Li, Chen, Feng, Guo,
  and {others}]{jiao_transformer_2023}
Licheng Jiao, Xin Zhang, Xu~Liu, Fang Liu, Shuyuan Yang, Wenping Ma, Lingling
  Li, Puhua Chen, Zhixi Feng, Yuwei Guo, and {others}.
\newblock Transformer meets remote sensing video detection and tracking: A
  comprehensive survey.
\newblock 16:\penalty0 1--45, 2023.

\bibitem[Khan et~al.(2022)Khan, Naseer, Hayat, Zamir, Khan, and
  Shah]{khan_transformers_2022}
Salman Khan, Muzammal Naseer, Munawar Hayat, Syed~Waqas Zamir, Fahad~Shahbaz
  Khan, and Mubarak Shah.
\newblock Transformers in vision: A survey.
\newblock 54\penalty0 (10):\penalty0 1--41, 2022.

\bibitem[Kobierski et~al.(2020)Kobierski, Lemanowicz, Wojew{\'o}dzki, and
  Kondratowicz-Maciejewska]{kobierski2020effect}
Miros{\l}aw Kobierski, Joanna Lemanowicz, Piotr Wojew{\'o}dzki, and Krystyna
  Kondratowicz-Maciejewska.
\newblock The effect of organic and conventional farming systems with different
  tillage on soil properties and enzymatic activity.
\newblock \emph{Agronomy}, 10\penalty0 (11):\penalty0 1809, 2020.

\bibitem[Lausch et~al.(2025)Lausch, Bumberger, Jung, Pause, Selsam, Zhou, and
  Herzog]{lausch2025monitoring}
Angela Lausch, Jan Bumberger, Andr{\'a}s Jung, Marion Pause, Peter Selsam, Tao
  Zhou, and Felix Herzog.
\newblock Monitoring agricultural land use intensity with remote sensing and
  traits.
\newblock \emph{Agriculture}, 15\penalty0 (21), 2025.

\bibitem[Leo et~al.(2023)Leo, Migliorati, Nguyen, and
  Grace]{leo_combining_2023}
Stephen Leo, Massimiliano De~Antoni Migliorati, Trung~H. Nguyen, and Peter~R.
  Grace.
\newblock Combining remote sensing-derived management zones and an
  auto-calibrated crop simulation model to determine optimal nitrogen
  fertilizer rates.
\newblock 205:\penalty0 103559, 2023.
\newblock {ISBN}: 0308-521X.

\bibitem[Li et~al.(2024)Li, Cai, Li, Kou, and Zhang]{li_review_2024}
Jiangyun Li, Yuanxiu Cai, Qing Li, Mingyin Kou, and Tianxiang Zhang.
\newblock A review of remote sensing image segmentation by deep learning
  methods.
\newblock 17\penalty0 (1):\penalty0 2328827, 2024.
\newblock {ISBN}: 1753-8947.

\bibitem[Lobert et~al.(2023)Lobert, Löw, Schwieder, Gocht, Schlund, Hostert,
  and Erasmi]{lobert_deep_2023}
Felix Lobert, Johannes Löw, Marcel Schwieder, Alexander Gocht, Michael
  Schlund, Patrick Hostert, and Stefan Erasmi.
\newblock A deep learning approach for deriving winter wheat phenology from
  optical and {SAR} time series at field level.
\newblock 298:\penalty0 113800, 2023.

\bibitem[Lobert et~al.(2025)Lobert, Schwieder, Alsleben, Broeg, Kowalski,
  Okujeni, Hostert, and Erasmi]{lobert_unveiling_2025}
Felix Lobert, Marcel Schwieder, Jonas Alsleben, Tom Broeg, Katja Kowalski,
  Akpona Okujeni, Patrick Hostert, and Stefan Erasmi.
\newblock Unveiling year-round cropland cover by soil-specific spectral
  unmixing of landsat and sentinel-2 time series.
\newblock 318:\penalty0 114594, 2025.
\newblock {ISBN}: 0034-4257.

\bibitem[Loshchilov and Hutter(2016)]{loshchilov_sgdr_2016}
Ilya Loshchilov and Frank Hutter.
\newblock Sgdr: Stochastic gradient descent with warm restarts.
\newblock 2016.

\bibitem[Loshchilov and Hutter(2017)]{loshchilov_decoupled_2017}
Ilya Loshchilov and Frank Hutter.
\newblock Decoupled weight decay regularization.
\newblock 2017.

\bibitem[Lu et~al.(2024)Lu, Guo, Zimmer-Dauphinee, Nieusma, Wang,
  {VanValkenburgh}, Wernke, and Huo]{lu_ai_2024}
Siqi Lu, Junlin Guo, James~R. Zimmer-Dauphinee, Jordan~M. Nieusma, Xiao Wang,
  Parker {VanValkenburgh}, Steven~A. Wernke, and Yuankai Huo.
\newblock {AI} foundation models in remote sensing: A survey, 2024.
\newblock URL \url{http://arxiv.org/abs/2408.03464}.

\bibitem[Mondelaers et~al.(2009)Mondelaers, Aertsens, and
  Van~Huylenbroeck]{mondelaers_meta-analysis_2009}
Koen Mondelaers, Joris Aertsens, and Guido Van~Huylenbroeck.
\newblock A meta-analysis of the differences in environmental impacts between
  organic and conventional farming.
\newblock 111\penalty0 (10):\penalty0 1098--1119, 2009.

\bibitem[Palaniappan and Annadurai(2018)]{palaniappan2018organic}
SP~Palaniappan and K~Annadurai.
\newblock \emph{Organic farming theory \& practice}.
\newblock Scientific publishers, 2018.

\bibitem[Pereira and dos Santos(2021)]{pereira_chessmix_2021}
Matheus~Barros Pereira and Jefersson~Alex dos Santos.
\newblock Chessmix: spatial context data augmentation for remote sensing
  semantic segmentation.
\newblock In \emph{2021 34th {SIBGRAPI} Conference on Graphics, Patterns and
  Images ({SIBGRAPI})}, pages 278--285. {IEEE}, 2021.

\bibitem[Pongratz et~al.(2018)Pongratz, Dolman, Don, Erb, Fuchs, Herold, Jones,
  Kuemmerle, Luyssaert, Meyfroidt, and {others}]{pongratz_models_2018}
Julia Pongratz, Han Dolman, Axel Don, Karl-Heinz Erb, Richard Fuchs, Martin
  Herold, Chris Jones, Tobias Kuemmerle, Sebastiaan Luyssaert, Patrick
  Meyfroidt, and {others}.
\newblock Models meet data: Challenges and opportunities in implementing land
  management in earth system models.
\newblock 24\penalty0 (4):\penalty0 1470--1487, 2018.

\bibitem[Radeloff et~al.(2024)Radeloff, Roy, Wulder, Anderson, Cook, Crawford,
  Friedl, Gao, Gorelick, Hansen, and {others}]{radeloff_need_2024}
Volker~C Radeloff, David~P Roy, Michael~A Wulder, Martha Anderson, Bruce Cook,
  Christopher~J Crawford, Mark Friedl, Feng Gao, Noel Gorelick, Matthew Hansen,
  and {others}.
\newblock Need and vision for global medium-resolution landsat and sentinel-2
  data products.
\newblock 300:\penalty0 113918, 2024.

\bibitem[Rolf et~al.(2024)Rolf, Klemmer, Robinson, and
  Kerner]{rolf_mission_2024}
Esther Rolf, Konstantin Klemmer, Caleb Robinson, and Hannah Kerner.
\newblock Mission critical--satellite data is a distinct modality in machine
  learning.
\newblock 2024.

\bibitem[Rufin et~al.(2020)Rufin, Frantz, Yan, and
  Hostert]{rufin_operational_2020}
Philippe Rufin, David Frantz, Lin Yan, and Patrick Hostert.
\newblock Operational coregistration of the sentinel-2a/b image archive using
  multitemporal landsat spectral averages.
\newblock 18\penalty0 (4):\penalty0 712--716, 2020.

\bibitem[Sanders et~al.(2025)Sanders, Brinkmann, Chmelikova, Ebertseder,
  Freibauer, Gottwald, Haub, Hauschild, Hoppe, Hülsbergen, and
  {others}]{sanders_benefits_2025}
Jürn Sanders, Jan Brinkmann, Lucie Chmelikova, Florian Ebertseder, Annette
  Freibauer, Frank Gottwald, Almut Haub, Michael Hauschild, Johanna Hoppe,
  Kurt-Jürgen Hülsbergen, and {others}.
\newblock Benefits of organic agriculture for environment and animal welfare in
  temperate climates.
\newblock pages 1--19, 2025.

\bibitem[Schuster et~al.(2023)Schuster, Hagn, Mittermayer, Maidl, Hülsbergen,
  Schuster, Hagn, Mittermayer, Maidl, and Hülsbergen]{schuster_using_2023}
Johannes Schuster, Ludwig Hagn, Martin Mittermayer, Franz-Xaver Maidl,
  Kurt-Jürgen Hülsbergen, Johannes Schuster, Ludwig Hagn, Martin Mittermayer,
  Franz-Xaver Maidl, and Kurt-Jürgen Hülsbergen.
\newblock Using remote and proximal sensing in organic agriculture to assess
  yield and environmental performance.
\newblock 13\penalty0 (7), 2023.
\newblock ISSN 2073-4395.
\newblock \doi{10.3390/agronomy13071868}.
\newblock URL \url{https://www.mdpi.com/2073-4395/13/7/1868}.
\newblock Company: Multidisciplinary Digital Publishing Institute Distributor:
  Multidisciplinary Digital Publishing Institute Institution: Multidisciplinary
  Digital Publishing Institute Label: Multidisciplinary Digital Publishing
  Institute.

\bibitem[Schwieder et~al.(2016)Schwieder, Leitão, da~Cunha~Bustamante,
  Ferreira, Rabe, and Hostert]{schwieder_mapping_2016}
Marcel Schwieder, Pedro~J Leitão, Mercedes~Maria da~Cunha~Bustamante,
  Laerte~Guimarães Ferreira, Andreas Rabe, and Patrick Hostert.
\newblock Mapping brazilian savanna vegetation gradients with landsat time
  series.
\newblock 52:\penalty0 361--370, 2016.

\bibitem[Shammi and Meng(2021)]{shammi_use_2021}
Sadia~Alam Shammi and Qingmin Meng.
\newblock Use time series {NDVI} and {EVI} to develop dynamic crop growth
  metrics for yield modeling.
\newblock 121:\penalty0 107124, 2021.
\newblock {ISBN}: 1470-160X.

\bibitem[Tarasiou et~al.(2023)Tarasiou, Chavez, and
  Zafeiriou]{tarasiou2023vits}
Michail Tarasiou, Erik Chavez, and Stefanos Zafeiriou.
\newblock Vits for sits: Vision transformers for satellite image time series.
\newblock In \emph{Proceedings of the IEEE/CVF Conference on Computer Vision
  and Pattern Recognition}, pages 10418--10428, 2023.

\bibitem[Tscharntke et~al.(2021)Tscharntke, Grass, Wanger, Westphal, and
  Batáry]{tscharntke_beyond_2021}
Teja Tscharntke, Ingo Grass, Thomas~C Wanger, Catrin Westphal, and Péter
  Batáry.
\newblock Beyond organic farming–harnessing biodiversity-friendly landscapes.
\newblock 36\penalty0 (10):\penalty0 919--930, 2021.

\bibitem[Turkoglu et~al.(2021)Turkoglu, D'Aronco, Perich, Liebisch, Streit,
  Schindler, and Wegner]{turkoglu_crop_2021}
Mehmet~Ozgur Turkoglu, Stefano D'Aronco, Gregor Perich, Frank Liebisch,
  Constantin Streit, Konrad Schindler, and Jan~Dirk Wegner.
\newblock Crop mapping from image time series: Deep learning with multi-scale
  label hierarchies.
\newblock 264:\penalty0 112603, 2021.
\newblock {ISBN}: 0034-4257.

\bibitem[Tóth and Kučas(2016)]{toth_spatial_2016}
Katalin Tóth and Andrius Kučas.
\newblock Spatial information in european agricultural data management.
  requirements and interoperability supported by a domain model.
\newblock 57:\penalty0 64--79, 2016.

\bibitem[Waldner and Diakogiannis(2020)]{waldner_deep_2020}
François Waldner and Foivos~I Diakogiannis.
\newblock Deep learning on edge: Extracting field boundaries from satellite
  images with a convolutional neural network.
\newblock 245:\penalty0 111741, 2020.

\bibitem[Wei et~al.(2019)Wei, Zhang, Wang, Wang, and
  Xu]{wei_multi-temporal_2019}
Sisi Wei, Hong Zhang, Chao Wang, Yuanyuan Wang, and Lu~Xu.
\newblock Multi-temporal {SAR} data large-scale crop mapping based on u-net
  model.
\newblock 11\penalty0 (1):\penalty0 68, 2019.

\bibitem[Weiss et~al.(2020)Weiss, Jacob, and Duveiller]{weiss_remote_2020}
Marie Weiss, Frédéric Jacob, and Grgory Duveiller.
\newblock Remote sensing for agricultural applications: A meta-review.
\newblock 236:\penalty0 111402, 2020.

\bibitem[Xie et~al.(2019)Xie, Dash, Huete, Jiang, Yin, Ding, Peng, Hall, Brown,
  Shi, and {others}]{xie_retrieval_2019}
Qiaoyun Xie, Jadu Dash, Alfredo Huete, Aihui Jiang, Gaofei Yin, Yanling Ding,
  Dailiang Peng, Christopher~C Hall, Luke Brown, Yue Shi, and {others}.
\newblock Retrieval of crop biophysical parameters from sentinel-2 remote
  sensing imagery.
\newblock 80:\penalty0 187--195, 2019.

\bibitem[Yuan et~al.(2022)Yuan, Lin, Liu, Hang, and
  Zhou]{yuan_sits-former_2022}
Yuan Yuan, Lei Lin, Qingshan Liu, Renlong Hang, and Zeng-Guang Zhou.
\newblock {SITS}-former: A pre-trained spatio-spectral-temporal representation
  model for sentinel-2 time series classification.
\newblock 106:\penalty0 102651, 2022.

\bibitem[Zhou et~al.(2025)Zhou, Ferdinand, van Wesemael, Dvorakova, Baret,
  Van~Oost, and van Wesemael]{zhou2025framework}
Yue Zhou, Manon~S Ferdinand, Jelle van Wesemael, Klara Dvorakova, Philippe~V
  Baret, Kristof Van~Oost, and Bas van Wesemael.
\newblock A framework for mapping conservation agricultural fields using
  optical and radar time series imagery.
\newblock \emph{Remote Sensing of Environment}, 328:\penalty0 114858, 2025.

\bibitem[Zhu et~al.(2017)Zhu, Tuia, Mou, Xia, Zhang, Xu, and
  Fraundorfer]{zhu_deep_2017}
Xiao~Xiang Zhu, Devis Tuia, Lichao Mou, Gui-Song Xia, Liangpei Zhang, Feng Xu,
  and Friedrich Fraundorfer.
\newblock Deep learning in remote sensing: A comprehensive review and list of
  resources.
\newblock 5\penalty0 (4):\penalty0 8--36, 2017.

\bibitem[Zhu et~al.(2015)Zhu, Wang, and Woodcock]{zhu_improvement_2015}
Zhe Zhu, Shixiong Wang, and Curtis~E Woodcock.
\newblock Improvement and expansion of the fmask algorithm: Cloud, cloud
  shadow, and snow detection for landsats 4–7, 8, and sentinel 2 images.
\newblock 159:\penalty0 269--277, 2015.

\end{thebibliography}

\appendix
\renewcommand{\thetable}{Appendix A\arabic{table}}
\setcounter{table}{0}

\begin{table}[htb]

    \centering

    \tabcolsep=0.11cm
    \scriptsize
     \begin{threeparttable}
      \caption{Data distribution of Train, Validation and Test data set, share relative to total area in respective data set}
    \label{table:DistributionComparison}
     \begin{tabular}{l|lll|lll|lll}
     \toprule
      \multicolumn{1}{c}{\:}
      & \multicolumn{3}{c}{\textbf{Training Data}}
      & \multicolumn{3}{c}{\textbf{Validation Data}}
      & \multicolumn{3}{c}{\textbf{Test Data}}\\
      \midrule \textbf{Crop\:type}   & $Area\:[ha]$ & $Share\:[\%]$   & $Org.\:[\%]$  & $Area\:[ha]$ & $Share\:[\%]$   & $Org. \:[\%]$ & $Area\:[ha]$ & $Share\:[\%]$   & $Org.\:[\%]$ \\
      \midrule
 
Winter wheat             & 26274.5 & 13.42 & 8.32  & 2159.2 & 8.78  & 6.86  & 5902.5  & 8.2   & 6.86  \\
Winter barley            & 12470.7 & 6.37  & 1.85  & 1476.8 & 6     & 1.01  & 3204.9  & 4.45  & 1.51  \\
Winter rye               & 2198.8  & 1.12  & 21.27 & 466.7  & 1.9   & 10.59 & 776.7   & 1.08  & 15.32 \\
Triticale                & 4168.2  & 2.13  & 10.82 & 859.7  & 3.49  & 5.84  & 1426    & 1.98  & 7.2   \\
Spring barley            & 4851.5  & 2.48  & 11.85 & 494.6  & 2.01  & 10.23 & 764.9   & 1.06  & 11.73 \\
Spring oat               & 1345.5  & 0.69  & 48.84 & 150.1  & 0.61  & 40.34 & 290.2   & 0.4   & 41.45 \\
Maize                    & 25137.2 & 12.84 & 2.85  & 2876.9 & 11.7  & 2.49  & 6464.2  & 8.98  & 2.45  \\
Potato                   & 1928.4  & 0.98  & 7.04  & 567.7  & 2.31  & 2.79  & 773.8   & 1.07  & 2.65  \\
Sugar beet               & 3840.1  & 1.96  & 4.23  & 273.6  & 1.11  & 1.85  & 914.4   & 1.27  & 3.28  \\
Winter rapeseed          & 4846.2  & 2.47  & 0.48  & 463    & 1.88  & 0.08  & 986     & 1.37  & 0.46  \\
Sunflower                & 323.1   & 0.17  & 30.87 & 8.1    & 0.03  & 21.99 & 37.6    & 0.05  & 32.69 \\
Cultivated grassland     & 6672.1  & 3.41  & 29.85 & 781.9  & 3.18  & 23.03 & 1575.8  & 2.19  & 26.81 \\
Permanent grassland      & 28714.5 & 14.66 & 11.97 & 4549.6 & 18.5  & 10.47 & 7939    & 11.03 & 10.2  \\
Vegetables               & 1036.4  & 0.53  & 21.67 & 49.2   & 0.2   & 32.68 & 214.5   & 0.3   & 13.64 \\
Peas                     & 831     & 0.42  & 28.7  & 72     & 0.29  & 21.71 & 182     & 0.25  & 19.38 \\
Broad bean               & 336.2   & 0.17  & 74.04 & 16.7   & 0.07  & 73.33 & 72.8    & 0.1   & 74.09 \\
Lupin                    & 52.5    & 0.03  & 63.44 & 4.3    & 0.02  & 74.32 & 14.8    & 0.02  & 70.23 \\
Soy                      & 794     & 0.41  & 28.9  & 55.1   & 0.22  & 26.39 & 157.1   & 0.22  & 32.62 \\
Hops                     & 762.4   & 0.39  & 1.83  & 32.2   & 0.13  & 0     & 959.2   & 1.33  & 0.31  \\
Grapevine                & 410.2   & 0.21  & 12.14 & 0.03      & 0     & 33.33 & 11.8    & 0.02  & 7.61  \\
Other agricultural areas & 660.5   & 0.34  & 23.35 & 69.5   & 0.28  & 22.22 & 194.9   & 0.27  & 22.41 \\
Orchard                  & 420.7   & 0.21  & 20    & 38.4   & 0.16  & 8.75  & 99.5    & 0.14  & 15.69 \\
Fallow land              & 2766.2  & 1.41  & 12.15 & 320.8  & 1.3   & 14.15 & 634     & 0.88  & 9.73  \\
Background               & 64967.7 & 33.18 & -  & 8810.7 & 35.82 & -  & 38403.3 & 53.34 & -                  
\end{tabular}
    \end{threeparttable}
\end{table}
\label{sec:sample:appendix}
\begin{table}[h]
\caption{Sample Distribution used for Random Forest Crop Type Model}
\centering
\footnotesize
\begin{tabular}{l|l}
Crop Type & Samples [n] \\
\toprule{}
Winter wheat & 97533 \\
Winter barley & 53350 \\
Winter rye & 11223 \\
Triticale & 20706 \\
Spring barley & 24089 \\
Spring oat & 8325 \\
Maize & 103435 \\
Potato & 8636 \\
Sugar beet & 10685 \\
Winter rapeseed & 17671 \\
Sunflower & 1222 \\
Cultivated grassland & 39948 \\
Permanent grassland & 232564 \\
Vegetables & 4587 \\
Peas & 3918 \\
Broad bean & 1269 \\
Lupin & 279 \\
Soy & 2719 \\
Hops & 3348 \\
Grapevine & 4932 \\
Other agricultural areas & 3807 \\
Orchard & 3524 \\
Fallow land & 26064 \\
Background & 19721
\end{tabular}
\end{table}
\begin{table}[h]
\caption{Sample Distribution used for Random Forest Management Form Model}
\centering
\footnotesize
\begin{tabular}{l|l}
Managemnt Form & Samples [n] \\
\toprule{}
Conventional & 617042 \\
Organic & 66792 \\
Background & 19721 
\end{tabular}
\end{table}

\begin{table*}[htb]
\footnotesize

\centering
\newcolumntype{Z}{>{\small}c}
\tabcolsep=0.15cm 
\caption{Classification results, F1-scores, Recall and Precision values for single crop types }
\resizebox{\textwidth}{!}{ %
\begin{tabular}{@{}p{2.35cm}ccc|ccc|ccc|ccc|ccc|ccc|ccc@{}}
\toprule
\textbf{} & \multicolumn{6}{c|}{\textbf{30 x 30}} & \multicolumn{6}{c|}{\textbf{10 x 10}} & \multicolumn{6}{c|}{\textbf{2 x 2}} & \multicolumn{3}{c}{\textbf{1 x 1}} \\
\midrule
\textbf{} & \multicolumn{3}{c|}{\textbf{Multitask}} & \multicolumn{3}{c|}{\textbf{Singletask}} & \multicolumn{3}{c|}{\textbf{Multitask}} & \multicolumn{3}{c|}{\textbf{Singletask}} & \multicolumn{3}{c|}{\textbf{Multitask}} & \multicolumn{3}{c|}{\textbf{Singletask}} & \multicolumn{3}{c}{\textbf{RF}} \\
\midrule
\textbf{Crop Types} & F1 & Re. & Pr . & F1 & Re. & Pr.  & F1 & Re. & Pr.  & F1 & Re. & Pr.  & F1 & Re. & Pr.  & F1 & Re. & Pr.  & F1 & Re. & Pr. \\
\midrule
Winter wheat & 0.94 & 0.94 & 0.95 & \textbf{0.95} & 0.94 & 0.95 & 0.94 & 0.94 & 0.94 & 0.94 & 0.94 & 0.94 & 0.93 & 0.93 & 0.94 & 0.92 & 0.92 & 0.93 & 0.80 & 0.72 & 0.90\\
Winter barley & 0.95 & 0.95 & 0.96 & \textbf{0.96} & 0.95 & 0.96 & 0.95 & 0.95 & 0.95 & 0.95 & 0.95 & 0.95 & 0.95 & 0.94 & 0.95 & 0.94 & 0.93 & 0.94 &0.79& 0.75& 0.84  \\
Winter rye & 0.83 & 0.86 & 0.81 & \textbf{0.84} & 0.85 & 0.83 & 0.81 & 0.81 & 0.81 & 0.82 & 0.84 & 0.80 & 0.80 & 0.84 & 0.77 & 0.78 & 0.79 & 0.76 & 0.20 & 0.40 & 0.13\\
Triticale & 0.83 & 0.82 & 0.83 & \textbf{0.84} & 0.83 & 0.86 & 0.81 & 0.82 & 0.81 & 0.81 & 0.83 & 0.79 & 0.79 & 0.80 & 0.79 & 0.77 & 0.79 & 0.75 & 0.40 & 0.58 & 0.31 \\
Spring barley & \textbf{0.89} & 0.88 & 0.89 & \textbf{0.89} & 0.89 & 0.90 & 0.87 & 0.85 & 0.89 & 0.87 & 0.87 & 0.88 & 0.87 & 0.84 & 0.90 & 0.85 & 0.86 & 0.85 & 0.71 & 0.63 & 0.81\\
Spring oat & \textbf{0.76} & 0.75 & 0.77 & \textbf{0.76} & 0.73 & 0.78 & 0.75 & 0.76 & 0.74 & \textbf{0.76} & 0.76 & 0.76 & 0.72 & 0.75 & 0.70 & 0.70 & 0.65 & 0.76 & 0.33 & 0.49 & 0.25 \\
Maize & \textbf{0.97} & 0.96 & 0.97 & \textbf{0.97} & 0.96 & 0.98 & \textbf{0.97} & 0.96 & 0.97 & \textbf{0.97} & 0.97 & 0.97 & \textbf{0.97} & 0.96 & 0.97 & 0.96 & 0.96 & 0.97 & 0.90 & 0.86 & 0.93\\
Potato & 0.95 & 0.96 & 0.94 & \textbf{0.96} & 0.96 & 0.95 & 0.95 & 0.94 & 0.95 & 0.95 & 0.95 & 0.95 & 0.95 & 0.95 & 0.94 & 0.94 & 0.95 & 0.93 & 0.85 & 0.87 & 0.83 \\
Sugar beet & \textbf{0.97} & 0.97 & 0.97 & \textbf{0.97} & 0.97 & 0.98 & \textbf{0.97} & 0.96 & 0.97 & \textbf{0.97} & 0.96 & 0.97 & 0.96 & 0.96 & 0.97 & 0.96 & 0.96 & 0.96 & 0.91 & 0.93 & 0.89\\
Winter rapeseed & \textbf{0.97} & 0.97 & 0.97 & \textbf{0.97} & 0.97 & 0.98 & \textbf{0.97} & 0.97 & 0.97 & \textbf{0.97} & 0.97 & 0.97 & \textbf{0.97} & 0.97 & 0.97 & 0.96 & 0.96 & 0.97 & 0.91 & 0.89 & 0.92\\
Sunflower & 0.81 & 0.79 & 0.83 & 0.86 & 0.86 & 0.85 & 0.82 & 0.88 & 0.77 & 0.84 & 0.88 & 0.81 & 0.79 & 0.80 & 0.79 & 0.78 & 0.80 & 0.75 & 0.36 & 0.80 & 0.23 \\
Cult. grassl. & \textbf{0.76} & 0.75 & 0.76 & \textbf{0.76} & 0.77 & 0.76 & 0.75 & 0.74 & 0.76 & \textbf{0.76} & 0.76 & 0.76 & 0.74 & 0.77 & 0.71 & 0.72 & 0.76 & 0.69 & 0.48 & 0.60 & 0.40\\
Perm. grassl. & 0.85 & 0.81 & 0.89 & 0.85 & 0.81 & 0.89 & \textbf{0.86} & 0.86 & 0.85 & \textbf{0.86} & 0.83 & 0.89 & \textbf{0.86} & 0.84 & 0.88 & 0.85 & 0.82 & 0.88 & 0.67 & 0.51 & 0.95 \\
Vegetables & \textbf{0.82} & 0.84 & 0.80 & \textbf{0.82} & 0.80 & 0.83 & 0.80 & 0.76 & 0.83 & 0.81 & 0.82 & 0.80 & 0.79 & 0.79 & 0.79 & 0.76 & 0.76 & 0.77 & 0.41 & 0.35 & 0.49\\
Peas & 0.85 & 0.86 & 0.84 & \textbf{0.87} & 0.87 & 0.87 & \textbf{0.87} & 0.88 & 0.86 & 0.86 & 0.87 & 0.85 & 0.85 & 0.87 & 0.84 & 0.82 & 0.79 & 0.87 & 0.56 & 0.66 & 0.49 \\
Broad bean & \textbf{0.86} & 0.92 & 0.81 & \textbf{0.86} & 0.88 & 0.84 & 0.85 & 0.86 & 0.84 & \textbf{0.86} & 0.90 & 0.82 & 0.84 & 0.84 & 0.84 & 0.83 & 0.87 & 0.79 & 0.61 & 0.89 & 0.46 \\
Lupin & 0.50 & 0.73 & 0.38 & \textbf{0.62} & 0.73 & 0.53 & 0.56 & 0.72 & 0.46 & 0.56 & 0.78 & 0.43 & 0.58 & 0.73 & 0.48 & 0.55 & 0.64 & 0.48 & 0.01 & 0.79 & 0.01\\
Soy & 0.89 & 0.92 & 0.86 & \textbf{0.91} & 0.95 & 0.88 & 0.89 & 0.92 & 0.85 & 0.90 & 0.93 & 0.87 & 0.88 & 0.94 & 0.82 & 0.87 & 0.92 & 0.83 & 0.63 & 0.86 & 0.49 \\
Hops & \textbf{0.97} & 0.97 & 0.97 & \textbf{0.97} & 0.96 & 0.98 & 0.96 & 0.96 & 0.96 & \textbf{0.97} & 0.97 & 0.96 & 0.96 & 0.97 & 0.96 & 0.96 & 0.97 & 0.96 & 0.78 & 0.99 & 0.65 \\
Grapevine & \textbf{0.75} & 0.67 & 0.85 & \textbf{0.75} & 0.68 & 0.82 & 0.71 & 0.61 & 0.86 & 0.74 & 0.66 & 0.84 & 0.74 & 0.69 & 0.80 & 0.67 & 0.63 & 0.72 & 0.13 & 0.07 & 0.69\\
Other agr. areas & 0.62 & 0.74 & 0.53 & \textbf{0.66} & 0.75 & 0.59 & 0.61 & 0.70 & 0.55 & 0.63 & 0.70 & 0.57 & 0.59 & 0.74 & 0.49 & 0.57 & 0.62 & 0.54 & 0.22 & 0.81 & 0.12 \\
Orchard & \textbf{0.34} & 0.49 & 0.26 & 0.33 & 0.52 & 0.24 & 0.31 & 0.49 & 0.23 & 0.32 & 0.53 & 0.23 & 0.27 & 0.59 & 0.17 & 0.26 & 0.57 & 0.17 & 0.02 & 0.08 & 0.01 \\
Fallow land & 0.59 & 0.61 & 0.58 & 0.59 & 0.63 & 0.55 & \textbf{0.60} & 0.61 & 0.59 & \textbf{0.60} & 0.64 & 0.56 & 0.57 & 0.57 & 0.57 & 0.55 & 0.54 & 0.57 & 0.30 & 0.24 & 0.40\\ 
Background & \textbf{0.96} & 0.97 & 0.95 & \textbf{0.96} & 0.97 & 0.95 & \textbf{0.96} & 0.96 & 0.96 & \textbf{0.96} & 0.97 & 0.96 & \textbf{0.96} & 0.96 & 0.96 & \textbf{0.96} & 0.96 & 0.95 & 0.88 & 0.99 & 0.80\\
\midrule\midrule
Mean F1 & 0.82 & 0.84 & 0.81 & \textbf{0.83} & 0.84 & 0.82 & 0.81 & 0.83 & 0.81 & 0.82 & 0.84 & 0.81 & 0.81 & 0.83 & 0.79 & 0.79 & 0.81 & 0.78 & 0.54 & 0.66 & 0.54 \\
\toprule
Overall accuracy & \textbf{0.93} &  & & \textbf{0.93} &  &  & \textbf{0.93} &  &  & \textbf{0.93} &  &  & \textbf{0.93} &  &  & 0.92 &  &  & 0.80 &  &  \\
\toprule
\end{tabular}}
\end{table*}

\end{document}